\documentclass[10pt,journal,compsoc]{IEEEtran}
\ifCLASSOPTIONcompsoc
  \usepackage[nocompress]{cite}
\else
  \usepackage{cite}
\fi

\usepackage{threeparttable}
\usepackage{amssymb}
\usepackage{comment}
\usepackage{xcolor}
\definecolor{darkblue}{RGB}{0,0,139}
\usepackage{xcolor}
\newcommand{\textbit}[1]{\textcolor{darkblue}{\textit{#1}}}

\usepackage{makecell}
\usepackage{amsmath,amsfonts,bm}
\usepackage{xspace}

\usepackage{booktabs}
\usepackage{tikz}
\usepackage[edges]{forest}
\definecolor{hiddendraw}{RGB}{20,68,106}
\definecolor{hidden-pink}{RGB}{255,245,247}
\definecolor{hidden-red}{RGB}{180,0,0}
\definecolor{output-black}{RGB}{0,0,0}
\definecolor{output-white}{RGB}{255,255,255}
\definecolor{myorange}{RGB}{255,208,153}
\definecolor{mygreen}{RGB}{166,207,152}
\definecolor{mygreen2}{RGB}{229,236,178}
\definecolor{myblue}{RGB}{83,198,240}
\definecolor{forestgreen}{RGB}{34,139,34}
\definecolor{darkgreen}{RGB}{24, 119, 24}

\usepackage[
    bookmarks=true,
    bookmarksnumbered=true,
    pdfpagemode={UseOutlines},
    plainpages=false,
    pdfpagelabels=true,
    colorlinks=true,
    linkcolor={darkblue},
    citecolor={darkblue},
    urlcolor={blue},
    pdftitle={Low-Rank Adaptation for Foundation Models: A Comprehensive Review},
    pdfsubject={Survey on Low-Rank Adaptation (LoRA) for Foundation Models},
    pdfauthor={Menglin Yang; Jialin Chen; Jinkai Tao; Yifei Zhang; Jiahong Liu; Jiasheng Zhang; Qiyao Ma; Harshit Verma; Regina Zhang; Min Zhou; Irwin King; Rex Ying},
    pdfkeywords={Foundation Models, LoRA, Parameter-Efficient Fine-Tuning, Survey}
]{hyperref}

\begin{document}
\title{Low-Rank Adaptation for Foundation Models: A Comprehensive Review}%

\author{Menglin Yang,
        Jialin Chen,
        Jinkai Tao,
        Yifei Zhang,
        Jiahong Liu,
        Jiasheng Zhang,
        Qiyao Ma,\\
        Harshit Verma,
        Regina Zhang,
        Min Zhou,
        Irwin King,
        Rex Ying
\IEEEcompsocitemizethanks{
\IEEEcompsocthanksitem Menglin Yang was with the Hong Kong University of Science and Technology (Guangzhou);
Jialin Chen, Harshit Verma and Rex Ying were with Yale University.
Yifei Zhang was with Nanyang Technological University.
Jinkai Tao was with Central University of Finance and Economics.
Jiahong Liu and Irwin King were with The Chinese University of Hong Kong; 
Jiasheng Zhang was with Xi’an University of Electronic Science and Technology. 
Qiyao Ma was with University of California, Davis; 
Regina Zhang was with The University of Cambridge. 
Min Zhou was with LOGS AI.
}
}

\markboth{}%
{Shell \MakeLowercase{\textit{et al.}}: Bare Advanced Demo of IEEEtran.cls for IEEE Computer Society Journals}
\IEEEtitleabstractindextext{%

\begin{abstract}
The rapid advancement of foundation models—large-scale neural networks trained on diverse, extensive datasets—has revolutionized artificial intelligence, enabling unprecedented advancements across domains such as natural language processing, computer vision, and scientific discovery. 
However, the substantial parameter count of these models, often reaching billions or trillions, poses significant challenges in adapting them to specific downstream tasks. 
Low-Rank Adaptation (LoRA) has emerged as a highly promising approach for mitigating these challenges, offering a parameter-efficient mechanism to fine-tune foundation models with minimal computational overhead. 
This survey provides the first comprehensive review of LoRA techniques beyond large language models to general foundation models, including recent technical foundations, emerging frontiers, and applications of low-rank adaptation across multiple domains. 
Finally, this survey discusses key challenges and future research directions in theoretical understanding, scalability, and robustness. 
This survey serves as a valuable resource for researchers and practitioners working with efficient foundation model adaptation.
Code is available at \url{https://github.com/marlin-codes/awesome-lora-adapter}
\end{abstract}

\begin{IEEEkeywords}
Foundation Model, Large Language Models, Low-Rank Adaptation, Parameter-Efficient Fine-Tuning, Multi-Task Learning
\end{IEEEkeywords}}

\maketitle

\IEEEdisplaynontitleabstractindextext

\IEEEpeerreviewmaketitle

\section{Introduction}
\label{sec:01.intro}
Foundation models represent a paradigm shift in artificial intelligence, wherein large-scale neural architectures, pre-trained on extensive and broad datasets, establish generalizable representational frameworks that can be adapted to a wide range of downstream applications~\cite{bommasani2021opportunities,zhou2023comprehensive}. 
These models span multiple domains, including natural language processing (e.g., GPT-3.5~\cite{brown2020language}, LLaMA~\cite{touvron2023llama}), computer vision (e.g., Swin Transformer~\cite{liu2021swin}, MAE~\cite{he2022masked}, SAM~\cite{kirillov2023segment}), speech processing (e.g., Wav2vec2~\cite{baevski2020wav2vec}, Whisper~\cite{radford2022robust}), multi-modal learning (e.g., Stable Diffusion~\cite{rombach2022high}, DALL·E 2~\cite{ramesh2022hierarchical}), and scientific applications (e.g., AlphaFold~\cite{jumper2021highly}, ChemBERTa~\cite{chithrananda2020chemberta}, ESM-2~\cite{lin2023evolutionary}). 

Foundation models are characterized by their unprecedented scale, with parameter counts reaching billions or even trillions, and exhibit emergent properties, capabilities that arise spontaneously without explicit training~\cite{bommasani2021opportunities}. 
These architectures have become fundamental building blocks of modern AI systems, enabling breakthrough performance across diverse domains~\cite{bommasani2021opportunities,zhou2023comprehensive}. 
While these models exhibit broad capabilities, task-specific optimization through fine-tuning remains essential for enhancing generalization~\cite{radford2021learning}, promoting algorithmic fairness~\cite{bolukbasi2016man}, enabling customization~\cite{hu2021lora}, and aligning with ethical and societal standards~\cite{bender2021dangers, weidinger2021ethical}.
However, their scale introduces significant computational challenges, particularly in the computational resources required for both training and fine-tuning~\cite{kaplan2020scaling}. Although traditional fine-tuning methods involving full parameter updates have demonstrated effectiveness across various tasks~\cite{liu2019roberta,wei2021finetuned}, their computational demands often render them impractical for foundation models~\cite{rajbhandari2020zero,han2024parameter}.

\begin{figure}[t]
\centering
\includegraphics[width=0.7\linewidth]{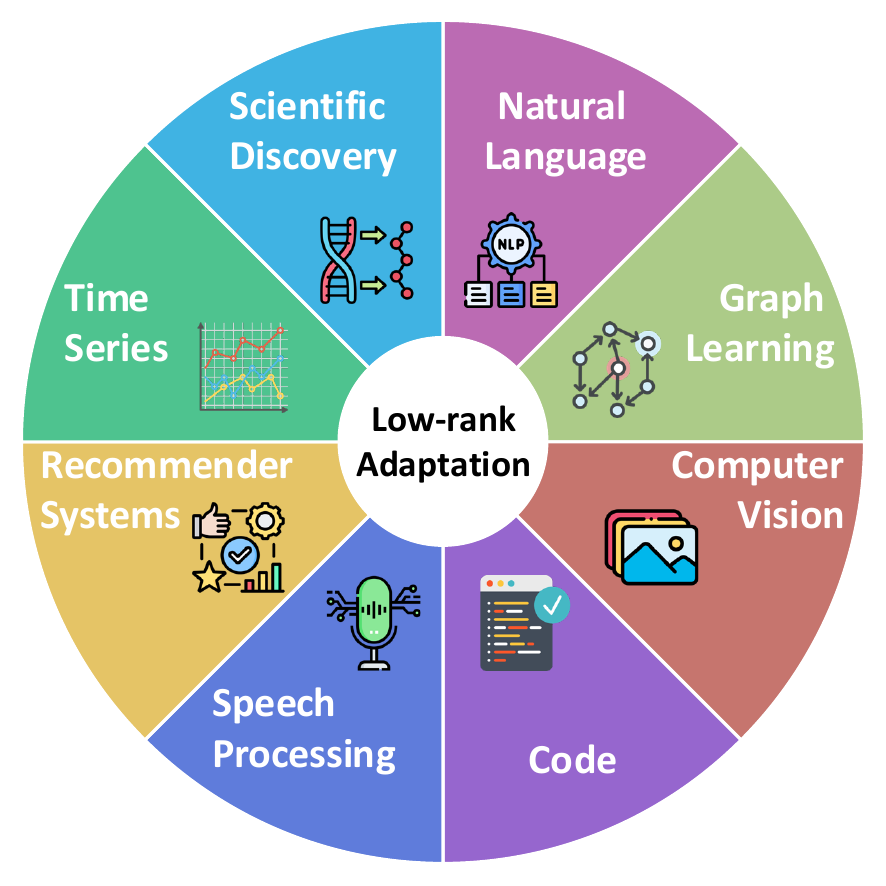}
\caption{LoRA with foundation models in diverse domains.}
\label{fig:intro_lora_domains}
\end{figure}

\begin{figure*}[t]
    \centering
    \includegraphics[width=1.0\linewidth]{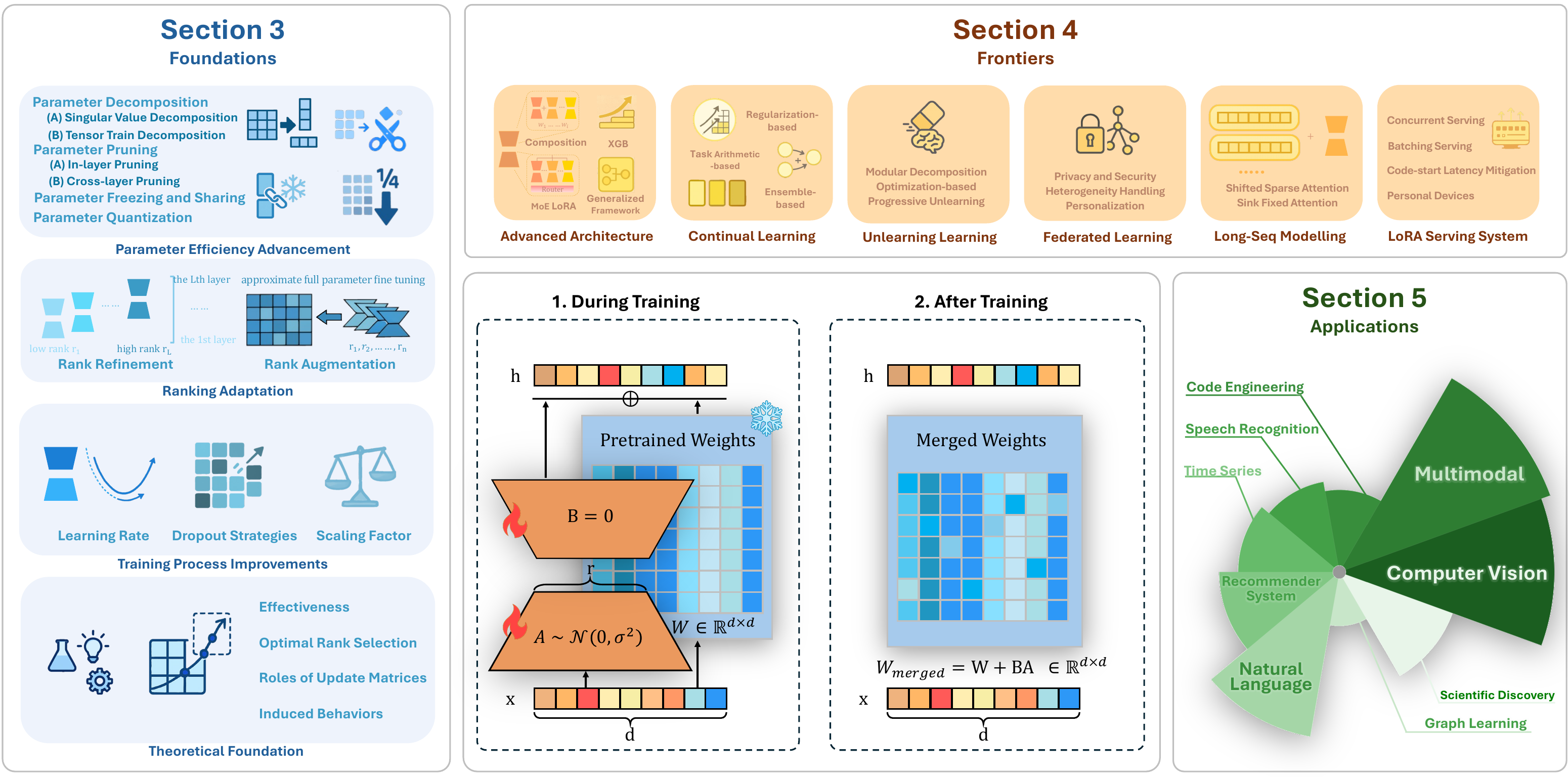}
    \caption{Overview of the LoRA landscape for foundation models, highlighting the core technical components, representative enhancements, and downstream applications covered in this survey.}
    \label{fig:intro_lora_frameworks}
\end{figure*}

Parameter-efficient fine-tuning (PEFT) methodologies have emerged as a solution to these computational challenges~\cite{houlsby2019parameter,hu2021lora,zaken2021bitfit,liu2022few,li2022blip,han2024parameter}. 
These approaches enable model adaptation by minimizing the number of trainable parameters, substantially reducing computational requirements without compromising task performance. Among these approaches, Low-Rank Adaptation (LoRA)~\cite{hu2021lora} and its variants have gained widespread attention due to their simplicity, empirical effectiveness, and broad applicability across diverse model architectures and domains, as shown in Figure~\ref{fig:intro_lora_domains}.

LoRA is grounded on two key insights: weight updates during fine-tuning often reside in a low-dimensional subspace~\cite{aghajanyan2020intrinsic,li2018measuring}, and task-specific adaptations can be effectively captured using low-rank matrices~\cite{hu2021lora}.
By optimizing these low-rank matrices per task while freezing the original model parameters, LoRA achieves efficient adaptation and enables the composition of multiple task-specific adaptations without increasing inference latency~\cite{lialin2023scaling,hu2021lora}.

\textbf{Contributions.} This survey provides, to the best of our knowledge, the first comprehensive review of LoRA-based techniques beyond the domain of Large Language Models (LLMs)~\cite{mao2024survey}, extending the analysis to the broader landscape of foundation models. Our key contributions are:

(1)  \textit{Systematic Analysis of Technical Foundations}: We provide a structured analysis of the recent technical advances of LoRA, including parameter efficiency strategies, rank adaptation mechanisms, training process improvements, and emerging theoretical perspectives.

(2) \textit{Extensive Investigation of Emerging Frontiers}: 
We explore emerging research frontiers, including advanced architectures incorporating multiple LoRA compositions and mixture-of-experts approaches, as well as methodologies for continual learning, unlearning, federated learning, long-sequence modeling, and efficient serving infrastructure.

(3) \textit{Comprehensive Review of Applications}: We present a comprehensive review of practical applications across diverse domains, including natural language processing, computer vision, speech recognition, scientific discovery, and specialized applications in code engineering, recommender systems, graph learning, and spatial-temporal forecasting.

This survey organizes existing LoRA research as illustrated in Figure~\ref{fig:intro_lora_frameworks}. It also highlights the critical challenges and future research directions in Section~\ref{sec:06.discussion}, with Figure~\ref{fig:structure_of_the_survey} presenting a detailed structure diagram of the survey. This survey offers a valuable resource for both researchers and practitioners in the field.

\section{Basics}
\label{sec:02.lora}
LoRA~\cite{hu2021lora} constitutes a substantial advancement in parameter-efficient fine-tuning (PEFT). Although originally developed for LLMs, subsequent research has demonstrated its effectiveness across a diverse set of foundation models.

The mathematical formulation of LoRA centers on constraining the {update matrix} $\Delta {W}$ to be low-rank during fine-tuning, as shown in Fig.~\ref{fig:intro_lora_frameworks}, which is implemented through matrix decomposition:
\begin{equation}
\Delta {W} = {B}{A},
\end{equation}
where ${B} \in \mathbb{R}^{d \times r}$, ${A} \in \mathbb{R}^{r \times k}$, and the rank $r \ll \min(d, k)$. 
By restricting $\Delta {W}$ to be low-rank, LoRA minimizes the number of parameters that need to be learned during the fine-tuning process, resulting in significant computational and storage efficiency.

\textbf{Parameter Initialization Strategies.}  
LoRA employs specific initialization strategies to ensure stable and efficient training. Matrix $A$ is typically initialized with values drawn from a random Gaussian distribution, while matrix $B$ is initialized with zeros, which ensures that at the start of training, $\Delta W = {B}{A}$ is effectively a zero matrix. 

\tikzstyle{leaf}=[draw=hiddendraw,
    rounded corners,minimum height=1em,
    fill=myblue!20,text opacity=1, align=center,
    fill opacity=.5, text=black, align=left,font=\scriptsize,
    inner xsep=3pt,
    inner ysep=1pt
    ]
\tikzstyle{middle}=[draw=hiddendraw,
    rounded corners,minimum height=1em,
    fill=output-white!40,text opacity=1, align=center,
    fill opacity=.5,  text=black,align=center,font=\scriptsize,
    inner xsep=3pt,
    inner ysep=1pt
    ]
\begin{figure*}[t]
\centering
\begin{forest}
  for tree={
  forked edges,
  grow=east,
  reversed=true,
  anchor=base west,
  parent anchor=east,
  child anchor=west,
  base=middle,
  font=\scriptsize,
  rectangle,
  line width=0.7pt,
  draw=output-black,
  rounded corners,align=left,
  minimum width=2em,
    s sep=5pt,
    inner xsep=3pt,
    inner ysep=2pt,
  },
  where level=1{text width=5em}{},
  where level=2{text width=6em,font=\scriptsize}{},
  where level=3{font=\scriptsize}{},
  where level=4{font=\scriptsize}{},
  where level=5{font=\scriptsize}{},
[ LoRA for Foundation Models , middle, align=center, rotate=90,anchor=north,edge=black,text width=11.0em
    [Foundations\\(\textsection\ref{sec:03.foundations}), middle, align=center, edge=output-black,text width=5.5em
        [Parameter Efficiency \\ Enhancement(\textsection\ref{subsec:parameter_efficiency_enhancement}), middle, text width=12.0em, align=center, edge=output-black
            [ {(1) Parameter Decomposition, (2) Parameter Pruning,\\
  (3) Parameter Freezing and Sharing, (4) Parameter Quantization}, 
              leaf, text width=25.0em, edge=output-black
            ]
        ]
        [Ranking Adaptation \\  (\textsection\ref{subsec:strategies}), middle, text width=12.0em, align=center, edge=output-black
            [ {(1)Rank Refinement, (2)Rank Augmentation},
              leaf, text width=25.0em, edge=output-black
            ]
        ]
        [Training Process \\ Improvements(\textsection\ref{subsec:training_process_enhancements}), middle, text width=12.0em, align=center, edge=output-black
            [ {(1)Learning Rate, (2)Dropout Strategies, (3)Scaling Factor},
              leaf, text width=25.0em, edge=output-black
            ]
        ]
        [Theoretical Foundations \\  (\textsection\ref{subsec:theoretical_basis}), middle, text width=12.0em, align=center, edge=output-black
            [ {(1)Effectiveness, (2)Optimal Rank Selection,\\(3)Roles of Update matrices, (4)Induced Behaviors}, 
              leaf, text width=25.0em, edge=output-black]
        ]
    ]
   [Frontiers\\(\textsection\ref{sec:04.frontiers}), middle, align=center, edge=output-black,text width=5.5em
        [Advanced Architecture \\  (\textsection\ref{subsec:advanced_architecture}), middle, text width=12.0em, align=center, edge=output-black
            [ {(1)LoRA Composition, (2)Generalized Framework,\\(3)Gradient Boosting LoRA, (4)MoE with LoRA},
              leaf, text width=25.0em, edge=output-black]
        ]
        [Continual Learning \\  (\textsection\ref{subsec:lora_continual_learning}), middle, text width=12.0em, align=center, edge=output-black
            [ {(1)Regularization-based, (2)Task Arithmetic-based,\\(3)Ensemble-based},
              leaf, text width=25.0em, edge=output-black]
        ]
        [Unlearning Learning \\  (\textsection\ref{subsec:lora_unlearning}), middle, text width=12.0em, align=center, edge=output-black
            [ {(1)Modular Decomposition, (2)Optimization-Based,\\(3)Progressive Unlearning},
              leaf, text width=25.0em, edge=output-black]
        ]
        [Federated Learning \\  (\textsection\ref{subsec:lora_federated_learning}), middle, text width=12.0em, align=center, edge=output-black
            [ {(1)Privacy and Security, (2)Computation Efficiency,\\(3)Heterogeneity Handling, (4)Personalization},
              leaf, text width=25.0em, edge=output-black]
        ]
        [Long Sequence Modeling \\  (\textsection\ref{subsec:lora_long_sequence_modeling}), middle, text width=12.0em, align=center, edge=output-black
            [ {(1)Shifted Sparse Attention, (2)Sink Fixed Attention},
              leaf, text width=25.0em, edge=output-black]
        ]
        [LoRA Serving Systems \\  (\textsection\ref{subsec:lora_serving_systems}), middle, text width=12.0em, align=center, edge=output-black
            [ {(1)Concurrent Serving, (2)Batching Serving,\\(3)Cold-start Latency Mitigation, (4)Personal Devices},
              leaf, text width=25.0em, edge=output-black]
        ]
    ]
   [Applications\\(\textsection\ref{sec:05.applications}), middle, align=center, edge=output-black,text width=5.5em
        [Language Tasks\\  (\textsection\ref{subsec:lora_language_tasks}), middle, text width=12.0em, align=center, edge=output-black
            [ {(1)NLU, QA, MT, Reasoning, NLG,\\(2)Multilingual Language and Dialects Processing,\\(3)Medical and Clinical Text Processing,etc.},
              leaf, text width=25.0em, edge=output-black]
        ]
        [Computer Vision\\  (\textsection\ref{subsec:lora_computer_vision}), middle, text width=12.0em, align=center, edge=output-black
            [ {(1)Visual Understanding, (2)Visual Generation},
              leaf, text width=25.0em, edge=output-black]
        ]
        [Speech Recognition\\  (\textsection\ref{subsec:lora_speech_recognition}), middle, text width=12.0em, align=center, edge=output-black
            [ {(1)Fake Audio Detection, (2)Multilingual ASR, (3)Low-Resource ASR},
              leaf, text width=25.0em, edge=output-black]
        ]
        [Code Engineering\\  (\textsection\ref{subsec:lora_code_engineering}), middle, text width=12.0em, align=center, edge=output-black
            [ {(1)Code Review and Analysis,\\(2)Code Generation and Summarization},
              leaf, text width=25.0em, edge=output-black]
        ]
        [Scientific Discovery\\  (\textsection\ref{subsec:lora_scientific_discovery}), middle, text width=12.0em, align=center, edge=output-black
            [ {(1)Protein Analysis, (2)Material Design},
              leaf, text width=25.0em, edge=output-black]
        ]
        [Recommender Systems\\  (\textsection\ref{subsec:lora_recommender_systems}), middle, text width=12.0em, align=center, edge=output-black
            [ {(1)CTR Prediction, (2)Sequential Recommendation},
              leaf, text width=25.0em, edge=output-black]
        ]
        [Graph Learning\\  (\textsection\ref{subsec:lora_graph_learning}), middle, text width=12.0em, align=center, edge=output-black
            [ {(1)Cross-Domain Graph Adaption,\\(2)Dynamic Knowledge Graph Update,\\(3)Equivariant Graph Adaption},
              leaf, text width=25.0em, edge=output-black]
        ]
        [Spatial-Temporal Forecasting\\  (\textsection\ref{subsec:lora_spatial-temporal_forecasting}), middle, text width=12.0em, align=center, edge=output-black
            [ {(1)Node-Specific Adaptation, (2)Multi-Channel Modeling,\\(3)Out-of-Domain Prediction},
              leaf, text width=25.0em, edge=output-black]
        ]
        [Multi Modal\\  (\textsection\ref{subsec:multi_modal}), middle, text width=12.0em, align=center, edge=output-black
            [ {(1)Language-Vision Learning, (2)Language-Audio Learning},
              leaf, text width=25.0em, edge=output-black]
        ]
    ]    
]
\end{forest}
\caption{Structure of LoRA for Foundation Models.}
\label{fig:structure_of_the_survey}
\end{figure*}
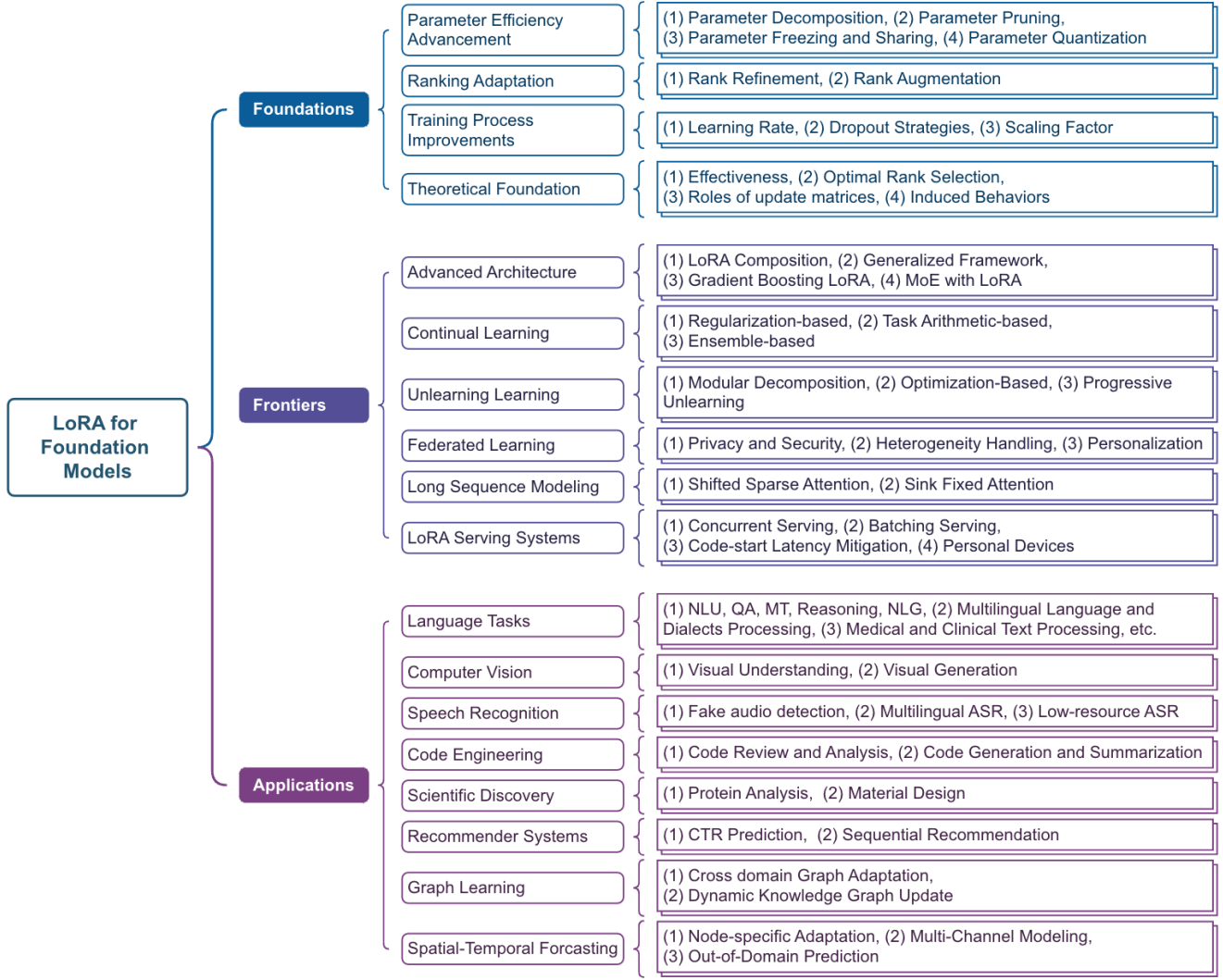

\textbf{Fine-tuning Process.}
In LoRA, the fine-tuning process follows these key principles:
\begin{itemize}
    \item The original pretrained weights ${W}_0$ are kept frozen and do not receive gradient updates during training.
    \item The low-rank matrices ${A}$ and ${B}$ contain the only trainable parameters, capturing task-specific adjustments.
    \item Both ${W}_0$ and $\Delta {W}$ are applied to the input vector ${x}$ separately and their outputs are combined.
    \item The output $\Delta {W} {x}$ is scaled by $ {\alpha}/{r}$. 
    \item The resulting output vectors are summed element-wise:
\end{itemize}
\begin{equation}
f{(x)} = {W}_0 {x} + \Delta {W} {x} = {W}_0 {x} + \frac{\alpha}{r}{B}{A} {x},
\end{equation}
where $\alpha/r$ is a scaling factor controlling the magnitude of the low-rank update.
When optimizing using Adam~\cite{kingma2014adam}, tuning the scaling factor $\alpha$ becomes roughly analogous to adjusting the learning rate~\cite{hu2021lora}, provided that the initialization is scaled appropriately. In practice, the value of $\alpha$ can be set based on the rank $r$, eliminating the need for extensive hyperparameter tuning.

\textbf{Advantages of LoRA over full fine-tuning}. 
LoRA offers several key advantages over full fine-tuning when applied to large foundation models: 

(1) \textit{Parameter Efficiency}. LoRA introduces a minimal set of trainable parameters through low-rank decomposition, typically reducing the number of task-specific parameters by several orders of magnitude. This is especially beneficial in resource-constrained and multi-task scenarios requiring multiple model adaptations. 

(2) \textit{Enhanced Training Efficiency}. Unlike conventional full fine-tuning, which updates all model parameters, LoRA optimizes only the low-rank matrices, substantially reducing computational costs and memory requirements, especially for models with billions of parameters. The reduced parameter space typically leads to faster convergence during training.

(3) \textit{No Additional Inference Latency}. LoRA does not introduce extra inference latency because the update matrix $\Delta W$ can be explicitly incorporated into the original frozen weights $W$. This integration ensures that the adapted model maintains efficiency during deployment and inference.

(4) \textit{Flexible Modular Adaptation}. LoRA enables the creation of lightweight, task-specific adapters that can be interchanged without modifying the base model architecture. This modularity facilitates efficient multi-task learning and task switching while minimizing storage requirements compared to maintaining separate model instances for each task. 

(5) \textit{Robust Knowledge Retention}. By preserving the pre-trained weights, LoRA effectively mitigates catastrophic forgetting, a common challenge in conventional fine-tuning. LoRA maintains the model's foundational knowledge while acquiring task-specific capabilities.

(6) \textit{Versatile Deployment}. The compact nature of LoRA adaptations facilitates efficient deployment and system integration. Multiple adaptations can be readily combined or alternated across different tasks or domains, offering enhanced flexibility compared to traditional fine-tuning approaches.

Through these advantages, LoRA enables efficient adaptation of foundation models while maintaining model performance and significantly reducing computational requirements.

\section{Foundations}
\label{sec:03.foundations}
In this section, we examine the fundamental technical aspects of LoRA across four critical dimensions: parameter efficiency enhancement, rank adaptation strategies, training process refinements, and theoretical foundations. These components constitute the technical foundation of LoRA's effectiveness.

\subsection{Parameter Efficiency Enhancement}
\label{subsec:parameter_efficiency_enhancement}

\begin{figure*}[!t]
    \centering
    \includegraphics[width=\linewidth]{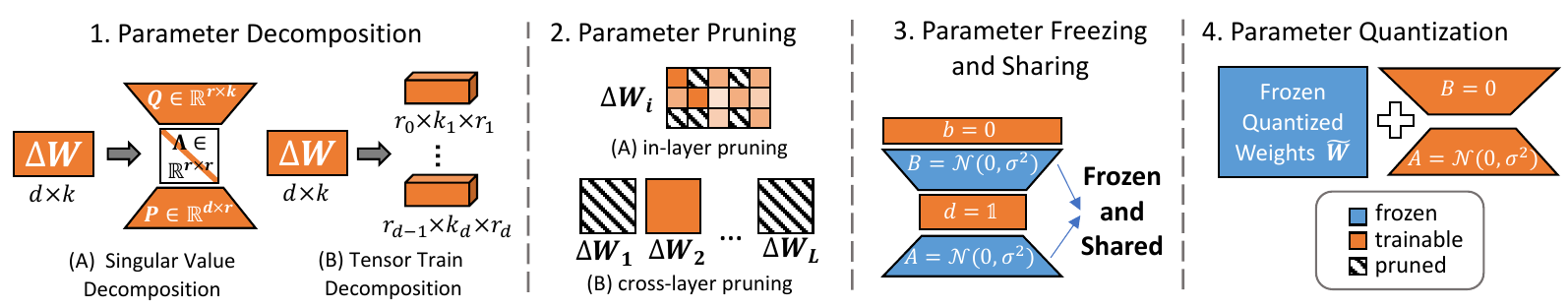}
    \caption{Illustration of parameter efficiency enhancement methods: decomposition, pruning, freezing and sharing, and quantization}
    \label{fig:parameter_efficiency_enhancement}
\end{figure*}

Despite the parameter efficiency gains achieved through LoRA with its project-down $A$ and project-up $B$ matrices, the method still requires a significant number of trainable parameters. 
For instance, applying LoRA to the LLaMA-2-70B model~\cite{touvron2023llama} necessitates updating over 16 million parameters~\cite{yang2024loretta}, surpassing the total parameter count of some BERT architectures~\cite{devlin2018bert}. 
Current research addresses this challenge through four primary approaches: parameter decomposition, pruning, freezing and sharing, and quantization. Fig.~\ref{fig:parameter_efficiency_enhancement} illustrates examples of these techniques.

\subsubsection{Parameter Decomposition} 
\label{subsec:parameter_decomposition}

Parameter decomposition methods achieve parameter efficiency by decomposing matrices in more compact forms while maintaining task performance. 
Beyond reducing trainable parameters, these methods also enable more granular control during fine-tuning. 
Current methodologies can be categorized into two principal approaches: update matrix decomposition~\cite{zhang2023adalora,qiang2024bilora,yang2024loretta,anjum2024tensor}, and pre-trained weight decomposition~\cite{liu2024dora}.

\begin{table}[!t]
\centering
\caption{Summary of Weight Decomposition Methods for LoRA}
\label{tab:weight_decomposition_methods}
\resizebox{0.5\textwidth}{!}{
\begin{tabular}{lll}
\toprule
\textbf{LoRAs} & \textbf{Method} & \textbf{Decomposition Components} \\
\midrule
\textbf{AdaLoRA} \cite{zhang2023adalora} & SVD & $\Delta W$: $\Delta W \leftarrow P\Lambda Q^\top$ \\
\textbf{BiLoRA} \cite{qiang2024bilora} & SVD & $\Delta W$: $\Delta W \leftarrow P\Lambda Q^\top$ \\
\textbf{LoRETTA}$_{rep}$ \cite{yang2024loretta} & Tensor Train & $\Delta W$: $\Delta W \leftarrow \text{TT}(B) \cdot \text{TT}(A)$ \\
\textbf{LoRETTA}$_{adp}$ \cite{yang2024loretta} & Tensor Train & $\Delta W$: $\Delta W \leftarrow \text{TT}(\Delta W)$ (series)\\
\textbf{TT-LoRA} \cite{anjum2024tensor} & Tensor Train & $\Delta W$: $\Delta W \leftarrow \text{TT}(\Delta W)$ (parallel) \\
\textbf{DoRA} \cite{liu2024dora} & Normalization & $W_0$: $W_0 \leftarrow \|W_0\|_c \frac{W_0}{\|W_0\|_c}$ \\
\bottomrule
\end{tabular}
}
\end{table}

\textbf{(1) Update Matrix Decomposition.} 
In update matrix decomposition approaches, two primary strategies have emerged: singular value decomposition (SVD)-based methods and tensor train (TT)-based decomposition.

\textit{(i) SVD-based Methods}. \textbit{AdaLoRA}~\cite{zhang2023adalora} parameterizes the updates weights $\Delta W$ in the form of SVD~\cite{wall2003singular}:
\begin{equation}
W = W_0 + \Delta W = W_0 + P\Lambda Q,
\end{equation}
where $P \in \mathbb{R}^{d \times r}$ and $Q \in \mathbb{R}^{r \times k}$ represent the left and right singular vectors of $\Delta W$, respectively, and the diagonal matrix $\Lambda \in \mathbb{R}^{r \times r}$ contains the singular values. 
AdaLoRA dynamically adjusts the rank of $\Delta W$ based on importance scoring, enabling adaptive parameter efficiency during fine-tuning.
Building upon this, \textbit{BiLoRA}~\cite{qiang2024bilora} extends this framework with bi-level optimization, separating singular vector and value training across different data subsets to mitigate overfitting.

\textit{(ii) TT-based Decomposition.} 
\textbit{LoRETTA}~\cite{yang2024loretta} takes a different approach by employing TT decomposition~\cite{oseledets2011tensor}, which represents a matrix into a series of low-rank, small, three-dimensional tensors, commonly referred to as cores. 
Given a matrix $W\in \mathbb{R}^{d\times k}$, it is first reshaped into a tensor $\mathcal{W} \in \mathbb{R}^{k_1 \times \cdots \times k_d}$, where $\prod_{i=1}^d k_i = d \times k$.
The TT representation of $\mathcal{W}$ can be formulated as:
\begin{equation}
   \text{TT}(\mathcal{W}) \leftarrow \prod_{i=1}^{d} \mathcal{C}_i,
\end{equation}
where $\mathcal{C}_i\in \mathbb{R}^{r_{i-1} \times k_i \times r_i}$ represents a core tensor, and $[r_0, \cdots, r_d]$ denotes TT rank with $r_0 = r_d =1$. This decomposition reduces the parameter count from $d \times k$ to $\sum_{{i=1}}^d r_{i-1}k_ir_i$.

\textbit{LoRETTA} introduces two variants: LoRETTA${adp}$ and LoRETTA${rep}$. LoRETTA${adp}$ employs tensorized adapters, inserting these lightweight modules after each attention and feed-forward sub-layer in the transformer blocks. LoRETTA$_{rep}$, on the other hand, reparameterizes the weight matrix with tensor factors, offering an even more compact PEFT approach. It updates the weights using two unbiased tensorized layers, further reducing the number of trainable parameters while maintaining comparable performance.

\textbit{TT-LoRA}~\cite{anjum2024tensor} applies this concept directly to the low-rank matrices in the original LoRA formulation. Note that TT-LoRA operates as a parallel adapter, directly modifying the update matrices in the original LoRA formulation. In contrast, LoRETTA$_{adp}$ functions as a series of adapters inserted into the pre-trained model architecture.

\textbf{(2) Pre-trained Weight Decomposition}. \textbit{DoRA}~\cite{liu2024dora} decomposes the pre-trained weight $W_0$ into magnitude and directional components by normalization method:
\begin{equation}
W_0 = m \frac{V}{\|V\|_c} = \|W_0\|_c \frac{W_0}{\|W_0\|_c},
\end{equation}
where $m \in \mathbb{R}^{1 \times k}$ is initialized as the magnitude vector $\|W_0\|_c$, $V \in \mathbb{R}^{d \times k}$ is initialized as $W_0$ and kept frozen, and $\|\cdot\|_c$ denotes the vector-wise norm of a matrix across each column.
During fine-tuning, the weight is adapted as:
\begin{equation}
W' = m \frac{W_0 + BA}{\|W_0 + BA\|_c},
\end{equation}
where $m$ becomes trainable and $BA$ represents the LoRA update to the directional component. This decomposition enables independent optimization of magnitude and direction during fine-tuning, leading to learning patterns that more closely resemble full fine-tuning.

Both approaches offer unique advantages in terms of parameter efficiency and fine-tuning flexibility. The \textit{update matrix decomposition} methods focus on decomposing the incremental updates applied during fine-tuning, while \textit{pre-trained weight decomposition} directly modifies the structure of the original model weights, where Table~\ref{tab:weight_decomposition_methods} provides a detailed comparison of these methods.

\subsubsection{Parameter Pruning} 
\label{sec:parameter_pruning}
Parameter pruning techniques focus on assessing the importance of different parameters within the LoRA matrices and removing those deemed less important. These methods can be categorized based on their pruning approaches: importance-based pruning, regularization-based pruning, and output-based pruning.  

\textbf{(1) Importance-based Pruning.} 
These methods evaluate parameter importance using multiple metrics. \textbit{SparseAdapter}~\cite{he2022sparseadapter} applies traditional network pruning techniques to LoRA parameters, assessing importance through parameter magnitude, gradient information, and sensitivity analysis. \textbit{RoseLoRA}~\cite{wang2024roselora} extends this concept by implementing sensitivity-based scoring for row/column pruning, enabling selective knowledge updates while preserving low-rank adaptation benefits.

\textbf{(2) Regularization-based Pruning.} 
Regularization-based pruning techniques induce sparsity through optimization constraints. \textbit{SoRA}~\cite{ding2023sparse} utilizes a gating mechanism between the down-projection and up-projection matrices of LoRA, utilizing proximal gradient descent with L1 regularization. This approach enables automatic sparsification during training, with zero-valued elements eliminated post-training.

\textbf{(3) Output-based Pruning.} 
Output-based methods evaluate LoRA parameters based on their layer-wise impact. \textbit{LoRA-drop}~\cite{zhou2024lora} evaluates the importance of LoRA modules by analyzing the distribution of $\|\Delta W_i x_i\|^2$ across different layers. The method retains individual LoRA modules for the most important layers while sharing a single LoRA across other layers deemed less critical. The importance score computation utilizes $\sum_{x \in D_s} \|\Delta W_i x_i\|^2$, where $D_s$ represents the sampled dataset.

\begin{table*}[!t]
\centering
\caption{Comparison of Parameter Efficiency Methods for Low-Rank Adaptation}
\resizebox{\textwidth}{!}{%
\begin{tabular}{llll}
\toprule
\textbf{Method} & \textbf{Strategy} & \textbf{Mechanism} & \textbf{Core Innovation} \\
\midrule
\multicolumn{4}{l}{\textit{Parameter Pruning}} \\
\textbf{SparseAdapter}~\cite{he2022sparseadapter} & Importance-based & Parameter scoring & Multi-criteria importance evaluation (magnitude, gradient, sensitivity) \\
\textbf{SoRA}~\cite{ding2023sparse} & Regularization-based & Gated sparsification & L1-regularized gating for adaptive sparsity \\
\textbf{LoRA-Drop}~\cite{zhou2024lora} & Output-based & Layer impact analysis & Dynamic pruning based on layer-wise output contributions \\
\midrule
\multicolumn{4}{l}{\textit{Parameter Freezing}} \\
\textbf{LoRA-FA}~\cite{zhang2023lora} & Selective freezing & Fixed feature extraction & Random initialization and freezing of matrix $A$ \\
\textbf{AsyLoRA}~\cite{zhu2024asymmetry} & Theoretical design & Orthogonal projection & Random orthogonal $A$ with theoretical guarantees \\
\midrule
\multicolumn{4}{l}{\textit{Parameter Sharing}} \\
\textbf{VeRA}~\cite{kopiczko2023vera} & Full sharing & Vector-based adaptation & Shared frozen matrices with trainable scaling vectors \\
\textbf{NOLA}~\cite{koohpayegani2023nola} & Basis sharing & Linear combination & Shared basis matrices with trainable coefficients \\
\textbf{Tied-LoRA}~\cite{renduchintala2023tied} & Flexible sharing & Layer-wise tying & Unified framework for cross-layer parameter sharing \\
\bottomrule
\end{tabular}%
}
\label{tab:parameter_pruning_freezing_sharing_comparison}
\end{table*}

\subsubsection{Parameter Freezing and Sharing}
\label{sec:parameter_sharing}
Parameter freezing and sharing techniques reduce trainable parameters through matrix-wise freezing and cross-layer parameter sharing. 

\textbf{(1) Matrix-wise Freezing.} Research has revealed asymmetric roles of matrices $A$ and $B$ in adaptation. \textbit{LoRA-FA}~\cite{zhang2023lora} demonstrates that freezing a randomly initialized matrix $A$ while only updating $B$ can maintain model performance. \textbit{Asymmetric LoRA}~\cite{zhu2024asymmetry} provides theoretical foundations for this approach, showing that $A$ primarily acts as a feature extractor while $B$ serves as a task-specific projector. This leads to an enhanced design using a frozen random orthogonal matrix for $A$, further reducing parameters while preserving performance.

\textbf{(2) Cross-layer Parameter Sharing.} Several methods explore parameter sharing across network layers. \textbit{VeRA}~\cite{kopiczko2023vera} shares frozen matrices $A$ and $B$ across layers, training only scaling vectors for adaptation. \textbit{NOLA}~\cite{koohpayegani2023nola} extends this concept by representing $A$ and $B$ as trainable linear combinations of shared frozen basis matrices. \textbit{Tied-LoRA}~\cite{renduchintala2023tied} implements layer-wise parameter tying while keeping the shared matrices trainable, offering a flexible framework that includes VeRA as a special case when the shared matrices are frozen.

Combined with parameter pruning techniques (Section~\ref{sec:parameter_pruning}), these methods enable a significant reduction in parameter count while maintaining adaptation effectiveness. Table~\ref{tab:parameter_pruning_freezing_sharing_comparison} provides a comprehensive comparison of these approaches.

\subsubsection{Parameter Quantization}
\label{subsec:quantization}
Quantization~\cite{hubara2017quantized,micikevicius2017mixed,banner2019post} optimizes neural network complexity through lower-precision numerical representations, substantially reducing storage and computational requirements. For a comprehensive quantization background, readers may refer to \cite{gholami2021survey}. In LoRA contexts, quantization approaches are characterized by two primary dimensions: {quantization timing} and {quantization techniques}.

\textbf{(1) Quantization Timing.} Quantization timing refers to when quantization occurs before, during, or after fine-tuning.

\textit{Pre-finetuning quantization}. {Pre-finetuning quantization} is that the pretrained weights are quantized prior to any LoRA-based adaptation.
For example, \textbit{QLoRA} \cite{dettmers2024qlora} employs a 4-bit NormalFloat (NF4) quantization method.
Similarly, \textbit{LoftQ} \cite{li2023loftq} improves upon this by addressing discrepancies introduced by quantizing high-precision weights. 
LoftQ jointly quantizes the pretrained model while optimizing low-rank initializations using an iterative algorithm.

\textit{During-fine-tuning quantization}. During-fine-tuning quantization applies quantization both before and throughout the fine-tuning process. Methods like \textbit{QA-LoRA}~\cite{xu2023qa} leverage group-wise quantization to adjust the precision dynamically during training, ensuring a more balanced interaction between low-rank updates and quantized weights. \textbit{L4Q}~\cite{jeon-etal-2025-l4q} merges the LoRA parameters $A$ and $B$ with the original weight matrix $W_0$ to form a new matrix, which is then quantized, enabling simultaneous optimization of both the quantization and LoRA parameters.

\textit{Post-fine-tuning quantization}. Post-fine-tuning quantization, such as in \textbit{LQER}~\cite{zhang2024lqer}, occurs after fine-tuning is completed, focusing on quantization primarily for inference. LQER utilizes a low-rank SVD-based decomposition to minimize quantization errors, ensuring that the quantized weights closely match the original high-precision weights.

\textbf{(2) Quantization Techniques.} Approaches for LoRA include uniform quantization, non-uniform quantization, and mixed-precision quantization.

\textit{Uniform Quantization}. Uniform quantization assigns the same bit-width across all weights, regardless of their distribution. 
\textbit{QA-LoRA}~\cite{xu2023qa} applies uniform quantization with group-wise refinement, balancing the precision trade-offs to optimize memory efficiency and adaptation. Remarkably, \textbit{LowRA}~\cite{zhou2025lowra} utilizes a data-free post-training quantization approach, reducing the quantization precision to below 2 bits through low-rank tensor optimization. However, uniform quantization may struggle with non-uniformly distributed weights, which is where non-uniform quantization becomes more effective. 

\textit{Non-uniform Quantization.} \textbit{QLoRA}~\cite{dettmers2024qlora} employs non-uniform quantization, designed specifically for weights with Gaussian-like distributions, which helps allocate more precision where it is needed most, near zero. 
This approach allows for a better representation of smaller weights, which dominate in pretrained models. 

\begin{table*}[htbp]
\centering
\caption{Comparison of Quantization Methods for LoRA}
\label{tab:quantization-comparison}
\resizebox{\textwidth}{!}{%
\begin{tabular}{lcccccccc}
\toprule
\textbf{Method} & \textbf{Timing} & \textbf{Target} & \textbf{Precision} & \textbf{Technique} & \textbf{Low-Rank} & \textbf{Optimization} & \textbf{Memory Focus} & \textbf{Dequant} \\
\midrule
\textbf{QLoRA}~\cite{dettmers2024qlora} & Pre-FT & Pretrained & 4 bit & NormalFloat & Standard LoRA & Separate & FT\&Inference & Partial \\
\textbf{QA-LoRA}~\cite{xu2023qa} & Pre \& During FT & Pretrained & 2, 3, 4 bit & Group-wise & Q-aware LoRA & Joint & FT\&Inference & None \\
\textbf{L4Q}~\cite{jeon-etal-2025-l4q} & Pre \& During FT & Pretrained & 2, 3, 4 bit & Uniform \& NormalFloat & Q-aware LoRA & Joint & FT\&Inference & None \\
\textbf{LowRA}~\cite{zhou2025lowra} & Pre-FT & Pretrained & $<$2 bit & Uniform \& Group-wise & Standard LoRA & Separate & FT & None\\
\textbf{LoftQ}~\cite{li2023loftq} & Pre-FT & Pretrained & mixed & Uniform \& NormalFloat & Q-aware LoRA & Joint & FT & Partial \\
\textbf{LQER}~\cite{zhang2024lqer} & Post-FT & Pretrained & mixed & Group-wise \& adaptive & SVD-based LR & Q-error min & Inference & None \\
\textbf{QDyLoRA}~\cite{rajabzadeh2024qdylora} & During FT & Pretrained & mixed & Rank sampling & Dynamic LoRA & Rank selection & FT & None \\
\textbf{LQ-LoRA}~\cite{guo2023lq} & Pre-FT & Pretrained & mixed & ILP \& data-aware & Q-aware LoRA & Joint & FT\&Inference & Partial \\
\bottomrule
\end{tabular}
}
\begin{tablenotes}
\small
\item FT = Fine-tuning, Q = Quantization, LR = Low-Rank, ILP = Integer Linear Programming
\end{tablenotes}
\end{table*}
\textit{Mixed-precision Quantization.} Mixed-precision quantization provides more flexibility by dynamically adjusting the bit-width based on the weight matrix or layer. Methods like \textbit{LoftQ}~\cite{li2023loftq} and \textbit{LQ-LoRA}~\cite{guo2023lq} optimize quantization across model components. \textbit{LoftQ} quantizes the residuals of the weight matrix and refines low-rank components using SVD, iteratively minimizing quantization errors. \textbit{LQ-LoRA} extends this approach by using integer linear programming to dynamically configure the bit-width for each weight matrix and incorporates a data-aware mechanism, utilizing an approximation of the Fisher information matrix to improve accuracy while minimizing quantization loss.

To summarize, pre-fine-tuning quantization approaches like QLoRA and LoftQ generally provide greater memory savings by freezing the pretrained weights, while post-fine-tuning methods like LQER focus more on refining precision for inference. 
In terms of quantization techniques, non-uniform and mixed-precision methods, as seen in QLoRA, LoftQ, and LQ-LoRA, demonstrate superior performance in low-bit scenarios by offering more flexible precision allocation based on weight distribution. 
The timing of quantization and the specific quantization technique both play critical roles in determining the balance between memory efficiency and model performance. Table \ref{tab:quantization-comparison} provides a comprehensive summary of the discussed quantization methods.

Overall, parameter efficiency enhancement techniques in LoRA have evolved through four complementary approaches: decomposition, pruning, freezing and sharing, and quantization, each offering unique trade-offs between model performance and resource utilization. Building upon these efficiency gains, we next explore how rank adaptation strategies further advance LoRA's capabilities.

\subsection{Rank Adaptation}
\label{subsec:strategies}
\begin{figure}[ht]
    \centering
    \includegraphics[width=0.8\linewidth]{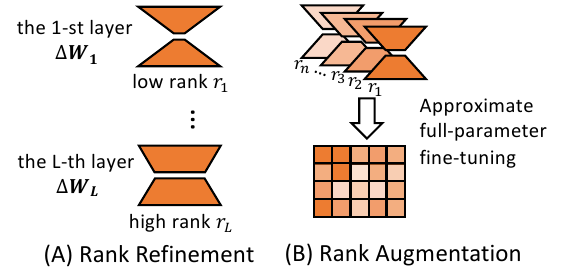}\vspace{-0.3cm}
    \caption{Illustration of rank refinement and augmentation methods.}
    \label{fig:rank_adjustment}
\end{figure}

Rank is a crucial parameter in LoRA, directly impacting the model adaptability and number of trainable parameters. 
The original LoRA method employs a fixed low rank across all layers, which may not be optimal for different downstream tasks and model architectures. To address this limitation, recent works have proposed various approaches to optimize rank allocation in LoRA, which can be broadly categorized into two main aspects: \textbf{rank refinement} and \textbf{rank augmentation}. Figure~\ref{fig:rank_adjustment} presents an illustration of these two methods.
\subsubsection{Rank Refinement}
Rank refinement methods aim to adaptively select the rank of LoRA modules during fine-tuning. The key insight is that different layers may require varying degrees of adaptation and thus benefit from different ranks. Rank refinement approaches can be grouped into three main types: adaptive allocation, heuristic strategies, and multi-rank training.

\textbf{(1) Adaptive Allocation.}  Adaptive allocation methods dynamically adjust the ranks of LoRA modules during training based on importance metrics derived from the data or model parameters.
\textbit{AdaLoRA}~\cite{zhang2023adalora} introduces an adaptive mechanism for rank allocation by parameterizing LoRA updates using SVD. It dynamically prunes singular values based on their magnitudes, allowing each layer to have a tailored rank while maintaining a global parameter budget. 

Similarly, \textbit{SoRA}~\cite{ding2023sparse} employs a learnable gating mechanism to control the effective rank of each LoRA module. The gates are optimized using proximal gradient descent with $\ell_1$ regularization to promote sparsity. This approach enables the automatic discovery of suitable ranks for different layers, improving parameter efficiency without manual tuning.

\textbf{(2) Heuristic Strategies.} Heuristic strategies allocate ranks based on predefined rules derived from prior knowledge or empirical observations.
\textbit{PRILoRA}~\cite{benedek2024prilora} proposes a deterministic strategy where the rank of LoRA modules increases linearly from lower to higher layers. Motivated by the observation that higher layers often require more adaptation in transfer learning, this heuristic allocates higher ranks to upper layers.

\textbf{(3) Multi-Rank Training.} Multi-rank training methods enable the model to perform well across a range of ranks, offering flexibility during inference.
\textbit{DyLoRA}~\cite{valipour2022dylora} trains LoRA modules across a spectrum of ranks simultaneously. In each training iteration, it samples ranks from a predefined distribution, allowing the model to learn to perform effectively across multiple ranks. This strategy enables adaptability during inference without requiring additional training, which is beneficial in deployment scenarios with varying computational constraints.

\subsubsection{Rank Augmentation}
Rank augmentation methods aim to achieve high-rank model updates through sequences of low-rank modifications, bridging the performance gap between LoRA and full-parameter fine-tuning. These methods can be categorized into three types: matrix merging-based, high-rank updating-based, and matrix resampling-based.

\textbf{(1) Matrix merging-based methods} increase the rank by merging low-rank update matrices. The key idea is that the sum of multiple low-rank matrices can approximate a higher-rank matrix, thereby enhancing the ability to capture complex patterns without incurring substantial computational overhead.

\textbit{ReLoRA}~\cite{lialin2023relora} and \textbit{COLA}~\cite{xia2024chain} introduce iterative optimization strategies to increase model rank. Both methods focus on periodically merging low-rank modules into pre-trained weights. ReLoRA employs an iterative framework where LoRA modules are periodically merged, resetting the optimizer and training new LoRA modules after each merge. Similarly, COLA, inspired by the Frank-Wolfe algorithm~\cite{lacoste2016convergence}, merges LoRA modules in a sequence, optimizing them to reduce residual errors and achieve high-rank expressiveness.

\textbit{MELoRA}~\cite{ren2024melora} introduces a parallelization approach to rank augmentation, which trains multiple small LoRA modules concurrently and concatenates their outputs to form a higher-rank adaptation. By assembling a mini-ensemble of low-rank adapters, MELoRA constructs an equivalent block-diagonal matrix that collectively has a higher rank.

\textbit{XGBLoRA}~\cite{yfzhang2024} integrates the Gradient Boosting (GB) framework into LoRA by combining Rank-1 boosters in sequence to refine the model's predictions, overcoming the dilemma between extremely low-rank adaptation and effectiveness.

\textbit{LoRA-LEGO}~\cite{zhao2025merging} performs rank-wise parameter clustering, grouping Minimal Semantic Units (MSUs) from different LoRAs into clusters and optimizes their scale using a dual reweighting strategy, achieving rank augmentation by merging low-rank matrices while maintaining computational efficiency.

\textbf{(2) High-Rank Updating-Based methods} enhance model capacity by increasing the rank of weight updates, typically using low-rank or non-trainable components, while maintaining computational and parameter efficiency.

\textbit{MoRA}~\cite{jiang2024mora} introduces a method for high-rank updating by using a square matrix, achieving a higher rank in the weight updates while maintaining the same number of trainable parameters. In addition, the method utilizes non-parameterized operators to reduce the input dimension and increase the output dimension for the square matrix, ensuring that the weight can still be merged back into the model.

\textbit{HiRA}~\cite{huang2025hira} utilizes the Hadamard product to construct a high-rank update matrix. Additionally, the authors propose HiLoRA, a combination of HiRA and LoRA, demonstrating that high-rank updates can provide greater capacity without additional inference overhead.

\textbit{RandLoRA}~\cite{albert2025randlora} constructs a high-rank update matrix by learning linear combinations of low-rank, non-trainable random matrices. During training, only the scaling coefficients of these matrices are optimized, enabling full-rank updates while preserving parameter efficiency. Additionally, it leverages sparse random matrices to further improve memory and computational efficiency, inspired by random projection techniques.

\textbf{(3) Resampling-based methods} achieve high-rank adaptations by dynamically resampling projection matrices during training, accumulating the effect of high-rank updates while using low-rank matrices at each step.

\textbit{FLoRA}~\cite{hao2024flora} reinterprets LoRA as a gradient compression and decompression mechanism. It periodically resamples the projection matrices used in LoRA modules during training. By changing these matrices at scheduled intervals, the method ensures that different subspaces are explored, effectively accumulating a higher-rank adaptation.

{\small
\begin{table}[!t] 
\centering 
\caption{Comparison of Rank Adjustment Methods for LoRA} 
\vspace{-10px}
\begin{tabular}{p{2.3cm}p{2.0cm}p{3.3cm}} 
\toprule 
\textbf{Adapter} & \textbf{Adjustment} & \textbf{Method} \\ 
\midrule 
\textbf{AdaLoRA}\cite{zhang2023adalora} & Refinement & Adaptive Allocation \\ 
\textbf{SoRA}\cite{ding2023sparse} & Refinement & Adaptive Allocation \\ 
\textbf{PRILoRA}\cite{benedek2024prilora} & Refinement & Heuristic Strategies \\ 
\textbf{DyLoRA}\cite{valipour2022dylora} & Refinement & Multi-Rank Training \\ 
\textbf{ReLoRA}\cite{lialin2023relora} & Augmentation & Matrix Merging-Based \\ 
\textbf{COLA}\cite{xia2024chain} & Augmentation & Matrix Merging-Based \\ 
\textbf{MELoRA}\cite{ren2024melora} & Augmentation & Matrix Merging-Based \\ 

\textbf{XGBLoRA}\cite{yfzhang2024} & Augmentation & Matrix Merging-Based \\ 

\textbf{LoRA-LEGO}\cite{zhao2025merging} & Augmentation & Matrix Merging-Based \\ 

\textbf{MoRA}\cite{jiang2024mora} & Augmentation & High-Rank Updating \\ 

\textbf{HiRA}\cite{huang2025hira} & Augmentation & High-Rank Updating \\ 

\textbf{RandLoRA}\cite{albert2025randlora} & Augmentation & High-Rank Updating \\ 

\textbf{FLoRA}\cite{hao2024flora} & Augmentation & Resampling-Based \\ 
\bottomrule 
\end{tabular} 
\vspace{-15px}
\label{tab:rank_adjustment} 
\end{table}}

In summary, rank adaptation strategies enhance LoRA adaptability by tailoring the rank of adaptation matrices to better suit the requirements of different layers and tasks. 
Table~\ref{tab:rank_adjustment} presents a detailed summary of rank refinement and augmentation.

\subsection{Training Process Improvements}
\label{subsec:training_process_enhancements}
While LoRA has demonstrated remarkable success in parameter-efficient fine-tuning, optimizing its training dynamics remains crucial for maximizing adaptation performance. 
In this section, we discuss recent advancements aimed at improving the training process, especially learning rates, dropout strategies, and scaling factors.

\textbf{Learning Rate.} In standard LoRA fine-tuning, a uniform learning rate is typically applied to both low-rank matrices $A$ and $B$. However, Hayou et al.~\cite{hayoulora+} observe that this practice leads to suboptimal performance, especially as the model width increases. The issue lies in the fact that updates to $A$ and $B$ contribute differently to the learning dynamics. 
To address this limitation, Hayou et al.~\cite{hayoulora+} propose \textbit{LoRA+}, a method that assigns different learning rates to matrices $A$ and $B$. Their theoretical analysis in the infinite-width limit reveals that for efficient learning, the magnitudes of feature updates from both $A$ and $B$ should be $\Theta(1)$. This necessitates scaling the learning rates such that $\eta_B = \Theta(1)$ and $\eta_A = \Theta(n^{-1})$, where $n$ denotes the model width. In practice, LoRA+ introduces a fixed ratio $\lambda = \eta_B / \eta_A > 1$, allowing practitioners to tune a single learning rate while automatically adjusting the other. 

\textbf{Dropout Strategies.} Despite the reduced number of trainable parameters in LoRA-based models, overfitting remains a concern, particularly when fine-tuning small or specialized datasets. Traditional dropout techniques may not suffice to mitigate overfitting in this context. Wang et al.~\cite{wang2024lora} highlight this vulnerability and propose a comprehensive framework to address it through dropout along three dimensions: dropping position, structural pattern, and compensation measure. The \textit{dropping position} specifies where the noise is introduced, such as in the attention logits, weights, or hidden representations. The \textit{structural pattern} defines the granularity of unit deactivation, encompassing element-wise, column-wise, or span-wise patterns. The \textit{compensation measure} aims to minimize the discrepancy between training and inference phases by incorporating techniques like normalized rescaling or Kullback-Leibler divergence loss. Building on this framework, the authors present \textbit{HiddenKey}~\cite{wang2024lora}, a dropout method that combines column-wise dropout of attention logits with element-wise dropout of hidden representations, supplemented by a KL divergence loss. 

\textbf{Scaling Factor.} In LoRA, a scaling factor $\gamma_r = {\alpha}/{r}$ is applied. However, as Kalajdzievski~\cite{kalajdzievski2023rank} points out, this scaling factor can cause gradient collapse when increasing the adapter rank, resulting in slowed learning and diminished performance for higher-rank adapters. To overcome this limitation, Kalajdzievski~\cite{kalajdzievski2023rank} proposes \textbit{rsLoRA}, which redefines the scaling factor to be $\gamma_r = {\alpha}/{\sqrt{r}}$. This adjustment ensures that the adapters are \textit{rank-stabilized}, meaning that both the forward and backward pass maintain stable magnitudes relative to the rank, even as it becomes large. Theoretically derived in the infinite-width limit, this scaling factor prevents gradient collapse, enabling stable learning across different adapter ranks.

By adapting learning rates to the distinct roles of LoRA matrices, mitigating overfitting through structured dropout, and preventing gradient collapse with rank-stabilized scaling, these methods enhance both the efficiency and effectiveness of LoRA fine-tuning. We next examine the theoretical foundations underlying LoRA's performance.

\subsection{Theoretical Foundations}
\label{subsec:theoretical_basis}
While the practical advantages of LoRA are evident, understanding its underlying principles from a theoretical perspective is crucial. This section addresses key questions regarding its effectiveness, optimal rank selection, the roles of update matrices, and the behavioral changes it induces.

\textbf{Q1: Why does LoRA work effectively?}
LoRA achieves competitive performance with full fine-tuning while updating only a small subset of parameters. This phenomenon can be understood through the Neural Tangent Kernel (NTK) theory\footnote{NTK describes the evolution of neural network gradients during training}. Malladi et al.~\cite{malladi2023kernel} show that LoRA approximately preserves the kernel of the original model during fine-tuning. Specifically, with high probability~\cite{malladi2023kernel}, 
\begin{equation}
\text{Pr}\left[\exists i, j \in [N], \left|\mathcal{K}_{\text{LoRA}}^{(\text{SGD})}(i, j) - \mathcal{K}^{(\text{SGD})}(i, j)\right| \geq c^2\epsilon\right] \leq \delta
\label{equ:ntk_approx}
\end{equation}
where $\mathcal{K}_{\text{LoRA}}^{(\text{SGD})}$ and $\mathcal{K}^{(\text{SGD})}$ are the kernels induced by LoRA and full fine-tuning respectively, $N$ is the number of examples in the dataset, $c$ is an upper bound on the L2 norms of gradients and inputs, $\epsilon$ is the approximation error, and $\delta$ is the probability bound given by $4N^2 \exp(-(\epsilon^2 - \epsilon^3)r/4)$, where $r$ is the rank used in LoRA.

Although LoRA restricts updates to a low-rank subspace, it effectively targets the gradients most responsible for significant transformations in network behavior based on Equation~(\ref{equ:ntk_approx}). By focusing on these critical gradients, LoRA preserves the model's ability to generalize, ensuring that the network remains sensitive to essential input variations while being highly parameter-efficient.

\textbf{Q2: How many ranks are required for optimal LoRA performance?}
The rank in LoRA fine-tuning is crucial for understanding the expressivity of adaptation and maintaining computational efficiency.

Zeng and Lee~\cite{zeng2024the} conducted a comprehensive study on the expressive power of LoRA across different architectures. 
(i) For fully connected neural networks, they show that LoRA can adapt any model $f$ to accurately represent a smaller target model $\tilde{f}$ if the LoRA rank $r$ satisfies:
\begin{equation}
r \geq (width\text{ of }f) \times \sqrt{\frac{depth\text{ of }\tilde{f}}{depth\text{ of }f}}.
\end{equation}
(ii) For Transformer networks, they demonstrate that any model can be adapted to a target model of the same size with rank-$(\text{embedding\_size}/{2})$ LoRA adapters. These findings provide a theoretical foundation for determining the minimum rank necessary for effective adaptation across different architectures.

Complementing this work, Jang et al.~\cite{jang2024lora} analyzed LoRA training in the NTK regime, yielding several key insights:
(i) They proved that full fine-tuning (without LoRA) admits a low-rank solution of rank $r\lesssim \sqrt{N}$, where $N$ is the number of training data points. 
(ii) Using LoRA with rank $r \gtrsim \sqrt{N}$ eliminates spurious local minima, facilitating efficient global minima discovery. This result suggests a lower bound for the LoRA rank to ensure optimization stability.
(iii) They provided generalization guarantees for LoRA-adapted models, demonstrating that the generalization error is bounded by $O({1}/{\sqrt{N}})$. This bound offers reassurance about the performance of LoRA-adapted models on unseen data.

These theoretical analyses offer valuable guidance for hyperparameter tuning in LoRA applications. 

\textbf{Q3: What are the roles of update matrices $A$ and $B$?}
Zhu et al.~\cite{zhu2024asymmetry} provide a comprehensive analysis of the distinct roles played by matrices $A$ and $B$ in LoRA. Their work reveals an inherent asymmetry in these matrices, which has important implications for fine-tuning efficiency and model generalization.

The authors~\cite{zhu2024asymmetry} demonstrate that $A$ primarily functions as a feature extractor from the input, while $B$ projects these features towards the desired output. This asymmetry suggests that fine-tuning $B$ alone can be more effective than fine-tuning $A$. Notably, their analysis shows that a randomly initialized $A$ can perform nearly as well as a fine-tuned one, challenging the conventional practice of updating both matrices. Building on this insight, Zhu et al. derive generalization bounds for different LoRA variants using an information-theoretic framework. When fine-tuning only $B$, the generalization bound takes the form:
\begin{equation}
|\text{gen}(\mu, B)| \leq \sqrt{\frac{2rq\sigma^2 \ln 2}{n}\sum_{i\in I} d^{(i)}_\text{out}}
\end{equation}
where $r$ is the rank, $q$ is the quantization bits, $\sigma$ relates to the sub-Gaussianity of the loss, $n$ is the sample size, and $d^{(i)}_\text{out}$ is the output dimension of the $i$-th layer. This bound is tighter compared to updating both $A$ and $B$, suggesting that freezing $A$ as a random orthogonal matrix and only updating $B$ could potentially enhance the generalization to unseen data.

These findings align with and extend the insights from previous questions, particularly the discussion on optimal rank selection. By focusing on updating $B$ alone, researchers can potentially achieve better generalization while further reducing the number of trainable parameters, thus enhancing both the efficiency and effectiveness of LoRA fine-tuning.

\textbf{Q4: What behavioral changes does LoRA induce in the model?}
Koubbi et al.~\cite{koubbi2024impactloraemergenceclusters} analyzed the dynamics of attention matrices, demonstrating that LoRA-induced low-rank modifications maintain short-term stability in token clustering while facilitating significant long-term divergence in learned representations.

LoRA updates attention matrices $(Q, K, V)$ with low-rank matrices $(\tilde{Q}, \tilde{K}, \tilde{V})$, introducing controlled perturbations:
\begin{equation}
\tilde{Q}= Q + Q_A Q_B^T, \quad \tilde{K} = K + K_A K_B^T, \quad \tilde{V} = V + V_A V_B^T
\end{equation}
Token dynamics under LoRA are described by:
\begin{equation}
\dot{x}_i(t) = \sum_{j=1}^n P_{ij}(t) V x_j(t),
\end{equation}
where attention weights $P_{ij}(t)$ are based on the softmax of the Query and Key matrices.

LoRA maintains short-term stability of token clustering, with the Wasserstein distance $W_2(\mu_t, \nu_t)$ between perturbed and unperturbed token distributions remaining bounded:
\begin{equation}
W_2(\mu_t, \nu_t)^2 \leq 2C_1(R_t)^2 \cdot e^{2C_t e^{3K_t}}.
\end{equation}
A key result is the identification of a phase transition, where tokens bifurcate into new clusters after a critical time $T^*(\delta)$, governed by the eigenvalue gap $\lambda_1 - |\lambda_2|$ of the Value matrix. This shows how LoRA fine-tunes models without catastrophic forgetting, preserving token structure early in training while allowing controlled divergence.

These theoretical foundations of LoRA show its effectiveness, from its competitive performance explained by the NTK theory to its ability to prevent catastrophic forgetting through controlled token dynamics. The insights into optimal rank selection and the asymmetry of update matrices offer practical guidelines for improvements.

\section{Frontiers}
\label{sec:04.frontiers}

Building upon the technical foundations discussed above, which establish the core components and mechanisms of LoRA, this section explores frontier developments that extend the capabilities of LoRA in novel directions. These frontier developments leverage and combine their fundamental principles to enable new functionalities, tackle more complex tasks, and address challenges in model adaptation. 

\subsection{Advanced Architecture}
\label{subsec:advanced_architecture}
While the original LoRA method significantly enhanced the efficiency of fine-tuning and demonstrated performance comparable to full fine-tuning, it had limitations in flexibility, generalization, and handling multiple diverse tasks simultaneously. To address these limitations, researchers have developed advanced LoRA architectures to further improve performance, parameter efficiency, and generalization ability. 

\subsubsection{LoRA Composition}
One major innovation in advanced LoRA architectures is the dynamic composition of multiple LoRA modules to enhance adaptability and generalization across diverse tasks. 

\textbf{Optimization-based Composition.} \textbit{LoRAHub}~\cite{huang2023lorahub} leverages CMA-ES~\cite{hansen1996adapting} gradient-free optimization to determine optimal coefficients for combining LoRA modules. Through few-shot learning, it autonomously selects and integrates modules for new tasks without requiring manual expertise or gradient computation. Similarly, \textbit{LoRA-Flow}~\cite{wang2024loraflow} introduces dynamic fusion weights to adjust the impact of different LoRAs at each generation step, determined by a fusion gate with minimal parameters and outperforming baselines that use static task-level fusion weights.

\textbf{Retrieval-based Composition.} \textbit{LoraRetriever}~\cite{zhao2024loraretriever} implements dynamic retrieval and composition of LoRA modules based on input prompts. It embeds task-specific LoRAs into a shared space using instruction fine-tuning on a subset of tasks, then retrieves relevant modules using cosine similarity. The framework supports both module fusion and mixture strategies while maintaining efficient batch processing. Similarly, by clustering and selecting task and domain data, \textbit{Adapters Selector}~\cite{tian-etal-2025-adapters} locates the appropriate adapter in an adapter index according to the detected domain and task. Alternatively, \textbit{ComLoRA}~\cite{huang2025comlora} associates a learnable embedding vector with each LoRA adapter, trains a selector based on input prompts for dynamic selection, and merges the chosen adapter into the base model during inference to preserve low latency.

\textbf{Batch-oriented Composition.} 
\textbit{FLORA}~\cite{wen2023batched} enables each example in a minibatch to utilize unique low-rank adaptation weights through efficient matrix operations. This design significantly improves throughput and reduces latency compared to traditional batched approaches, particularly beneficial when serving diverse user requests in production environments.

\begin{figure}[!t]
    \centering
    \includegraphics[width=0.8\linewidth]{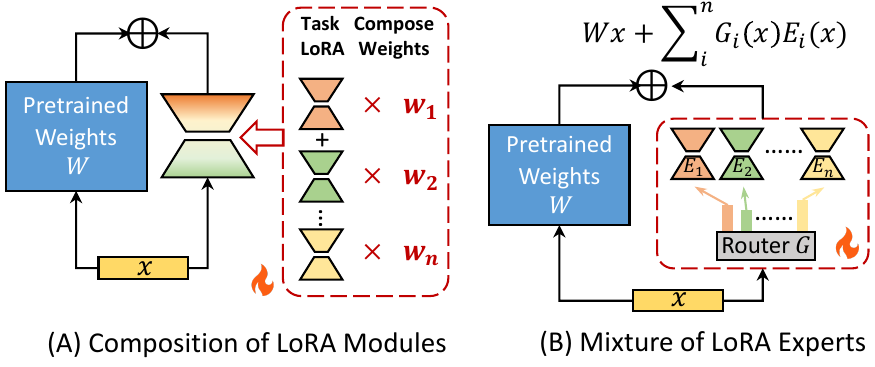}\vspace{-0.3cm}
    \caption{Illustration of LoRA composition and LoRA mixture of experts}
    \label{fig:lora_composition}
\end{figure}

By enabling models to select and combine multiple LoRA modules based on the task or input, these methods overcome the limitations of standard LoRA in handling diverse tasks and improve overall performance.

\subsubsection{Generalized Framework}
Another advancement involves extending the LoRA architecture itself to capture both task-specific and general features more effectively. 

\textbf{Dual-branch Framework.} 
\textbit{Hydra}~\cite{kim2023hydra} presents a generalized formulation by integrating both parallel and sequential LoRA branches within the model. 
The parallel branch learns task-specific features, similar to standard LoRA, while the sequential branch linearly combines pre-trained features. This dual-branch design enables Hydra to capture both task-specific adaptations and leverage general pre-trained knowledge, offering a comprehensive adaptation mechanism that improves performance across tasks.

\textbf{Multi-PEFT United Framework.} \textbit{GLoRA}~\cite{chavan2023one} further generalizes LoRA by unifying various parameter-efficient fine-tuning methods. It introduces trainable support tensors to scale and shift weights, features, and biases, effectively subsuming methods such as LoRA, adapter tuning, and prompt tuning within a single framework. GLoRA employs evolutionary search to determine optimal layer-wise configurations of these tensors, which can take scalar, vector, or low-rank matrix forms. Through structural re-parameterization, GLoRA incurs no additional inference cost while providing greater flexibility than previous PEFT methods.


These generalized architectures enhance the expressive power of LoRA by incorporating additional mechanisms for capturing diverse features and facilitating more effective fine-tuning across tasks.

{\small
\begin{table}[!t]
\centering
\caption{Comparison of Different LoRA Composition Methods}
\begin{tabular}{p{1.9cm}p{6.1cm}}
\toprule
\textbf{Method} & \textbf{Key Features and Mechanism} \\
\midrule
\textbf{LoRAHub}~\cite{huang2023lorahub} & Optimization-based composition using CMA-ES for gradient-free weight optimization; enables few-shot adaptation without manual expertise. \\
\midrule
\textbf{LoraRetriever}~\cite{zhao2024loraretriever} & Retrieval-based composition; supports both fusion and mixture strategies with efficient batch processing. \\
\midrule
\textbf{Adapters Selector}~\cite{tian-etal-2025-adapters} & Retrieval-based composition; dynamically selects adapters during inference. \\
\midrule
\textbf{ComLoRA}~\cite{huang2025comlora} & Retrieval-based composition; embeds LoRA adapters with learnable vectors and uses a selector to pick the winner. \\
\midrule
\textbf{LoRA-Flow}~\cite{wang2024loraflow} & Dynamic fusion with fusion gates to adjust LoRA impact in generative tasks. \\ \midrule
\textbf{FLORA}~\cite{wen2023batched} & Batch-oriented composition; achieves 3x throughput improvement at low ranks. \\
\midrule
\textbf{Hydra}~\cite{kim2023hydra} & Dual-branch design; combines parallel (new features) and sequential (pre-trained features) adaptations. \\
\midrule
\textbf{GLoRA}~\cite{chavan2023one} & Multi-PEFT unified framework with trainable tensors for weights/features/bias. \\ \midrule
\textbf{XGBLoRA}~\cite{yfzhang2024} & Gradient boosting framework with rank-1 LoRA adapters; combines multiple weak learners. \\
\bottomrule
\end{tabular}
\label{tab:lora_composition_comparison}
\end{table}
}

\subsubsection{Gradient Boosting with LoRA}
Gradient Boosting with LoRA (GBLoRA) combines weak learners through iterative LoRA module training to minimize residual errors. After $T$ boosting iterations, the finetuned model is expressed as:
\begin{equation}
\mathcal{M}^{T}(x) = \mathcal{M}_0 + \sum_{t=1}^T \mathcal{B}_{\Delta W^{(t)}}(x).
\end{equation}
with cumulative model updates:
\begin{equation}
\vspace{-3px}
\mathcal{M}^{(t)}(x) = \mathcal{M}^{(t-1)}(x) + \eta \mathcal{B}_{\Delta W^{(t)}}(x).
\end{equation}
where $\eta$ controls the contribution of each LoRA booster $\mathcal{B}_{\Delta W^{(t)}}(x)$. The weak learner principle enables GBLoRA to achieve strong performance with low-rank updates. \textbit{XGBLoRA}~\cite{yfzhang2024} established convergence guarantees and expressiveness bounds, demonstrating how increased boosting iterations can compensate for lower ranks. This framework unifies various matrix merging methods like ReLoRA~\cite{lialin2023relora}, COLA~\cite{xia2024chain}, and MeLoRA~\cite{ren2024melora} within the GB paradigm.

\subsubsection{Mixture of Experts with LoRA}
Another important advancement in LoRA architectures is combining LoRA with Mixture of Experts (MoE), where multiple ``expert" sub-networks specialize in different input patterns~\cite{shazeer2017outrageously}. A gating mechanism routes inputs to the most appropriate experts, allowing the model to handle a wide range of tasks efficiently~\cite{cai2024survey}. Given the input $x$, the MoE model computes 
\begin{equation}
    y = \sum_i^n G_i{(x)} E_i(x)
\end{equation}
where $y$ is the output, $G_i$ is a gating function, $E_i$ is an expert, and $n$ is the number of experts.

By integrating LoRA with MoE, models learn multiple pairs of low-rank matrices (LoRA experts) instead of a single pair, with a router determining the weights or selection of experts based on inputs. During fine-tuning, the pre-trained LLM weights remain fixed, while the LoRA experts and the router are trained, leveraging the parameter efficiency of LoRA and the specialization capabilities of MoE. The typical framework is illustrated in Figure~\ref{fig:lora_composition} (B).

Research on LoRA-MoE methods can be broadly categorized into three groups based on their primary objectives: (1) enhancing performance and parameter efficiency, (2) preserving knowledge during fine-tuning, and (3) adapting to multi-task learning. While these categories highlight different focuses, many approaches address multiple objectives simultaneously.

\textbf{(1) Efficiency-oriented Design.}
Methods in this category aim to match full fine-tuning performance with minimal parameter overhead.

Zadouri et al.~\cite{zadouri2023pushing} introduce \textit{MoV} and \textit{MoLoRA}, which achieve full fine-tuning parity while updating fewer than 1\% of parameters and improving generalization to unseen tasks. These methods treat (IA)$^3$ vectors and LoRA adapters as experts, respectively, and employ soft merging—where all experts contribute to the output weighted by router probabilities.

Building on this, Luo et al.~\cite{luo2024moelora} propose \textit{MoELoRA}, which integrates multiple LoRA experts within a Mixture-of-Experts (MoE) framework. It uses top-$k$ routing and a load-balancing loss~\cite{fedus2022switch} to avoid expert collapse, while contrastive learning among experts mitigates random routing issues~\cite{zuo2021taming}.

However, the fixed number of LoRA experts—for example, in MoELoRA~\cite{luo2024moelora}—lacks flexibility and can be redundant due to representation collapse or overfitting in the routing policy~\cite{chen2023sparse}.
To address this, Gao et al.~\cite{gao2024higher} introduce \textit{MoLA}, a layer-wise expert allocation strategy that dynamically assigns LoRA experts across Transformer layers via top-$k$ routing. Its sparse activation not only improves parameter efficiency but also supports continual learning by preserving knowledge from prior tasks.

More recently, \textit{MoKA}~\cite{yu-etal-2025-moka} replaces LoRA’s low-rank decomposition with Kronecker products within a sparse MoE architecture, compressing parameters while boosting performance. \textit{RepLoRA}~\cite{truong2025replora} reparameterizes LoRA matrices using a lightweight MLP within an MoE framework, reducing the sample complexity for accurate low-rank estimation from exponential to polynomial. Finally, \textit{GOAT}~\cite{fan2025make} closes the gap with full fine-tuning by adaptively integrating SVD-structured priors and aligning LoRA-MoE optimization with full fine-tuned MoE—enhancing performance without additional computational cost.

\textbf{(2) Memory-based Adaptation.}
These approaches focus on preventing catastrophic forgetting during adaptation.
Two notable approaches, \textit{LoRAMoE}~\cite{dou2024loramoe} and \textit{MoRAL}~\cite{yang2024moral}, address the challenge of knowledge retention while adapting LLMs to new tasks or domains.

\textbit{LoRAMoE}~\cite{dou2024loramoe} integrates multiple LoRA experts via a router network with a localized balancing constraint, allowing experts to focus on world knowledge for downstream tasks. It uses a top-$k$ routing strategy to balance world knowledge and task performance. \textbit{MoRAL}~\cite{yang2024moral} combines question-answer pairs from unstructured text with MoE's multitasking and LoRA's efficiency. It uses soft routing, where experts contribute based on router probabilities, adapting to new domains while preventing catastrophic forgetting.

{\small
\begin{table}[!t]
\setlength\tabcolsep{1pt}
    \fontsize{7}{7}\selectfont
\centering
\caption{Comparison of MoE-LoRA Methods}
\vspace{-10px}
\begin{tabular}{p{2.3cm}p{1.8cm}p{4cm}}
\toprule
\textbf{Method} & \textbf{Routing} & \textbf{Key Feature} \\ \midrule
\textbf{MoV/MoLoRA}~\cite{zadouri2023pushing} & Soft routing &  All experts contribute with weights \\
\textbf{MoELoRA}~\cite{luo2024moelora} & Top-k & Contrastive between experts \\
\textbf{MoLA}~\cite{gao2024higher} & Top-k & Layer-wise expert distribution \\
\textbf{LoRAMoE}~\cite{dou2024loramoe} & Top-k & Localized balancing for knowledge \\
\textbf{MoRAL}~\cite{yang2024moral} & Soft routing & Lifelong learning framework \\
\textbf{MOELoRA}~\cite{liu2024moe} & Task-based & Task identifier conditioning \\
\textbf{MoCLE}~\cite{gou2023mixture} & Cluster-based & Instruction cluster routing \\
\textbf{LLaVA-MoLE}~\cite{chen2024llava} & Token-level & Top-1 sparse expert selection \\
\bottomrule
\end{tabular}
\label{tab:moe_lora_comparison}
\vspace{-15px}
\end{table}
}

\textbf{(3) Task-based Integration.}
These methods tackle domain specificity and task interference challenges. Domain specificity arises when models trained on general-purpose data lack the specialized knowledge required for specific domains like medicine~\cite{liu2024moe} or finance~\cite{feng2024mixture}. Task interference occurs when multiple tasks and their datasets compete during training, leading to degraded performance across tasks~\cite{chen2024llava,gou2023mixture,feng2024mixture}. 

\textit{To address domain specificity}, Liu et al.~\cite{liu2024moe} propose \textit{MOELoRA} for multi-task medical applications, employing multiple LoRA experts with a task-aware gating function that routes inputs based on task identity—enabling task-specific adaptation while sharing a common knowledge base.  
Building on this paradigm, Feng et al.~\cite{feng2024mixture} introduce \textit{MOA}, an end-to-end parameter-efficient method that first trains task-specific LoRAs and then combines them via sequence-level routing conditioned on domain metadata, facilitating flexible composition of domain-specialized adapters.  
Further advancing dynamic specialization, Buehler et al.~\cite{buehler2024x} present \textit{X-LoRA}, which adopts a layer-wise token-level MoE strategy: starting from pretrained LoRAs, it dynamically mixes adapted layers using gating networks driven by hidden states, enabling novel adapter combinations and strong performance in scientific domains.

\textit{To solve task interference}, 
\textbit{MoCLE}~\cite{gou2023mixture} is an MoE architecture that activates task-specific parameters based on instruction clusters, using a cluster-conditional routing strategy and a universal expert to enhance generalization. 
\textbit{LLaVA-MoLE}~\cite{chen2024llava} proposes a sparse MoE design with multiple LoRA experts and a token-level routing strategy, allowing adaptive routing for tokens from different domains to address data conflicts. \textbit{HMoRA}~\cite{liao2025hmora} improves multi-task learning with a hierarchical hybrid routing strategy and auxiliary loss function, enhancing task distinction and token-context capture. \textbit{MoSLD}~\cite{zhao-etal-2025-mosld} shares the upper projection matrix among experts and applies a dropout strategy to enhance generalization across tasks. \textbit{SMoRA}~\cite{zhao2025rankexpertsinglerankedmixture} uses dynamic rank-wise activation, treating each LoRA rank as an expert to improve knowledge sharing and mitigate task conflicts in multi-task scenarios.

In addition, Tian et al.~\cite{tian2024hydralora} propose \textbit{HydraLoRA}, an asymmetric LoRA architecture that challenges the conventional symmetric expert structure in MoE-based approaches. 
Through empirical analysis, they discovered that in multi-task settings, matrix $A$ parameters from different LoRA heads tend to converge while matrix $B$ parameters remain distinct. Building on this observation, HydraLoRA introduces an architecture with a shared matrix $A$ across all tasks and multiple task-specific $B$ matrices, employing a trainable MoE router to automatically identify intrinsic components within the training data. 

By employing various routing strategies and expert designs, these methods enable efficient adaptation to multiple tasks or domains while mitigating interference and maintaining task-specific performance. 
The integration of MoE with LoRA has demonstrated promising results in enhancing performance, preserving knowledge, and facilitating multi-task adaptation across various domains.

\subsection{LoRA for Continual Learning}
\label{subsec:lora_continual_learning}
The parameter-efficient nature of LoRA allows for incrementally updating models on new tasks while mitigating catastrophic forgetting~\cite{wang2023orthogonal,wistuba2023continual}. Several key advantages motivate the use of LoRA for continual learning (CL): (1) reduced computational costs compared to full fine-tuning, (2) natural isolation of task-specific knowledge, and (3) flexible combination of task-specific adaptations. 
Existing LoRA-based continual learning methods can be broadly categorized into three approaches: regularization-based methods, task arithmetic-based methods, and ensemble-based techniques.

\textbf{Regularization-based} approaches leverage parameter constraints on LoRA updates as the primary mechanism to prevent catastrophic forgetting, focusing on preserving critical model parameters.
\textbit{O-LoRA}~\cite{wang2023orthogonal} addresses this by constraining new task updates to be orthogonal to previous tasks' subspaces, allowing incremental learning while preserving earlier tasks. \textbit{GORP}~\cite{wang-etal-2025-continual} combines full and low-rank parameters in a unified low-rank gradient subspace, mitigating forgetting and improving efficiency. \textbit{Online-LoRA}~\cite{wei2024online} enables task-free online continual learning for Vision Transformers, using weight regularization and real-time loss monitoring to maintain performance across changing data. \textbit{TreeLoRA}~\cite{qian2025treelora} optimizes task structures using hierarchical gradient similarity and sparse updates, enhancing performance in continual learning.

\textbf{Task arithmetic-based} approaches leverage task vector arithmetic on LoRA parameters. Chitale et al.~\cite{chitale2023task} apply arithmetic operations on LoRA parameters to combine knowledge from multiple tasks. Their method trains separate LoRA modules for each task and then uses task vector addition to create a task-agnostic model. A key insight is that LoRA parameters create semantic ``task vectors'' in weight space that can be manipulated algebraically. 

\textbf{Ensemble-based} works maintain and combine multiple task-specific LoRA modules.
\textbit{CoLoR}~\cite{wistuba2023continual} maintains separate LoRA modules for each task and uses an unsupervised approach to select the appropriate module at inference time. It trains task-specific LoRA modules sequentially and combines them using prototype-based task identification. This allows for the isolation of task knowledge while enabling flexible combinations. \textbit{AM-LoRA}~\cite{liu2024learning} uses multiple task-specific LoRA modules combined with an attention mechanism to integrate knowledge from different tasks. The attention-based mixing strategy enables adaptive knowledge integration while preventing catastrophic forgetting between tasks.

While these approaches demonstrate the potential of LoRA for continual learning, several challenges remain. The orthogonality constraint in O-LoRA may be overly restrictive for tasks with overlapping knowledge. Task arithmetic assumes tasks can be linearly combined, which may not hold for all scenarios. Ensemble methods face challenges in task identification and scaling to many tasks.

\subsection{LoRA for Unlearning}
\label{subsec:lora_unlearning}
LoRA facilitates the targeted removal of specific knowledge from foundation models without necessitating extensive retraining. This section categorizes and examines methodologies employing LoRA for unlearning, focusing on three primary categories: modular decomposition methods, optimization-based approaches, and a progressive unlearning pipeline.

\textbf{Modular Decomposition Methods}. These methods focus on decomposing and modularizing model components to support unlearning. Gao et al.~\cite{gao2024practical} introduce an orthogonal LoRA mechanism that ensures parameter disentanglement during successive unlearning processes. 
This design ensures that unlearning requests can be processed consecutively without causing interference with retained knowledge. 
Chen and Yang~\cite{chen2023unlearn} propose adding efficient unlearning LoRA layers, employing a selective teacher-student objective to guide the model in "forgetting" specific data. Additionally, Lizzo and Heck~\cite{lizzo2024unlearn} introduce UNLEARN, where LoRA layers are adapted to identify and isolate targeted knowledge in low-dimensional subspaces.

\textbf{Optimization-Based Approaches.}
These methods rely on optimization techniques to selectively remove or suppress specific knowledge from foundation models.  
Cha et al.~\cite{cha2024towards} propose a data-driven LoRA initialization strategy that weights low-rank updates by Fisher information, prioritizing parameters most critical for erasing targeted knowledge. In contrast, Gundavarapu et al.~\cite{gundavarapu2024machine} employ gradient ascent with low-rank LoRA updates to unlearn harmful or unwanted information, directly steering model outputs away from undesirable behaviors.

\textbf{Sequential Pipeline Strategies}. These approaches implement structured, multi-step procedures for systematic unlearning. Liu et al.~\cite{liu2024towards} leverage LoRA to negate specific harmful knowledge in a structured two-stage process. The first stage focuses on identifying harmful content, while the second stage applies LoRA to suppress and neutralize such knowledge without affecting other learned information. This methodical approach ensures systematic removal of unwanted information while preserving the model's general capabilities.

\subsection{LoRA for Federated Learning}
\label{subsec:lora_federated_learning}
In an era of heightened data privacy concerns, Federated Learning (FL) offers a promising approach to leverage collective knowledge while maintaining robust protection of individual data. The integration of LoRA into Federated Foundation Models (FFM) has made foundation models more accessible to resource-constrained devices, particularly in edge computing scenarios, potentially revolutionizing IoT and mobile applications. The combination of federated instruction tuning and value alignment with LoRA creates a powerful synergy that addresses several critical challenges in distributed machine learning.

\textbf{Privacy and Security.} Privacy protection is paramount in federated learning. \textbit{FedIT}~\cite{zhang2024towards} and \textbit{FFA-LoRA}~\cite{sun2024improving} establish frameworks that combine federated learning with instruction tuning for LLMs. These systems implement FedAvg privacy-preserving mechanisms, keeping instruction data on local devices while transmitting only encrypted LoRA parameters to the central server. Taking a different approach, \textbit{PrivateLoRA}~\cite{wang2023privatelora} exchanges only activations between the central cloud and edge devices to preserve data locality. Advancing security further, Huang et al.~\cite{huang2024fast} integrate model slicing with trusted execution environments (TEEs), employing server-side TEEs for later model layers and sparsification parameter fine-tuning (SPF) with LoRA to achieve both security and performance without requiring client-side TEEs.

\textbf{Computational Efficiency.} Despite significant advances in privacy and security, FL faces a fundamental trade-off between model expressiveness and computational efficiency. 
Research shows that to guarantee the ability to fit any target model, the rank of LoRA must meet a lower bound that scales with embedding size~\cite{zeng2024the} as discussed in Section~\ref{subsec:theoretical_basis}. 
However, implementing high ranks incurs significant communication and computation costs, especially for resource-constrained devices~\cite{he2020fedml, lim2020federated, horvoth2022natural}. To mitigate this, \textit{FedGBA}~\cite{yfzhang2024} combines ensemble learning with rank-1 LoRA updates for expressive yet efficient federated fine-tuning. 
\textit{FFA-LoRA}~\cite{sun2024improving} reduces overhead by fixing randomly initialized non-zero matrices and fine-tuning only zero-initialized ones. 
More recently, \textit{FedEx-LoRA}~\cite{singhal-etal-2025-fedex} achieves exact model updates with minimal cost by injecting a residual error term into the frozen pretrained weights, preserving both accuracy and efficiency.

\textbf{Heterogeneity Handling.} A key challenge in FL is managing heterogeneous data, devices, and models across clients, particularly in non-independently and identically distributed (non-IID) scenarios. 
\textit{SLoRA}~\cite{babakniya2023slora} tackles this via data-driven LoRA initialization. 
\textit{FedLoRA}~\cite{wang2024flora} aggregates client adapters of varying ranks through a stacking-based method. 
\textit{pFedLoRA}~\cite{yi2023fedlora} introduces a lightweight adapter to facilitate iterative global-local knowledge exchange during heterogeneous local training. 
Wagner et al.~\cite{wagner2024personalized} enhance fine-tuning robustness via trust-weighted gradient aggregation. 
\textit{HetLoRA}~\cite{cho2023heterogeneous} combines high- and low-rank adaptations to prevent overfitting and enable communication-efficient fine-tuning on resource-constrained devices. 
Recently, \textit{LoRA-A$^2$}~\cite{koo-etal-2025-towards} alternates between freezing LoRA modules and adaptively selecting ranks to mitigate aggregation inconsistency caused by data heterogeneity, ensuring stable performance under high heterogeneity.

\textbf{Personalization.} Personalization in FL adapts the global model to individual clients using techniques such as local fine-tuning, LoRA adaptation, or hybrid approaches, balancing performance, privacy, and efficiency based on each client's data and resource constraints.
\textbit{FedHLT} and \textbit{FedLFC}~\cite{guo2024fedlfc} combine low-rank adaptation with language-family clustering to address FFM challenges of high communication costs and parameter interference, achieving improved performance with reduced overhead compared to baselines. 
\textbit{PER-PCS}~\cite{tan2024personalized} allows users to safely share and collaboratively assemble personalized LoRA pieces for LLMs, reducing computation costs while maintaining high performance.
\textbit{FDLoRA}~\cite{xu2024dofit} employs dual LoRA modules—one for capturing global knowledge and another for client-specific adaptation—enabling efficient personalization of large language models while reducing communication and computation costs.
\textbit{pFedMixF}~\cite{zhang2025pfedmxf} addresses non-IID data, sequential tasks, and varying client capabilities through mixture-based frequency aggregation, balancing personalization and generalization while mitigating catastrophic forgetting in federated environments.

The integration of LoRA with FFM advances distributed machine learning by providing privacy-preserving model adaptation while balancing efficiency, heterogeneity, and personalization. Combining privacy mechanisms, efficient parameter transmission, and adaptive techniques, LoRA makes FL practical for resource-constrained environments without compromising model security or expressiveness. As edge computing and IoT evolve, the synergy between LoRA and FFM promises to advance distributed machine learning by enabling efficient, secure, and personalized model deployment in privacy-sensitive domains.

\subsection{LoRA for Long Sequence Modeling}
\label{subsec:lora_long_sequence_modeling}
The ability to process long sequences is crucial for many tasks across various domains handled by foundation models~\cite{Chen2023LongLoRA,Zhang2024SinkLoRA,yang2023longqlora}. However, standard foundation models are typically constrained by their maximum context length due to the quadratic computational complexity of self-attention with respect to sequence length. 
To address this limitation, several LoRA-based techniques have been proposed to extend the context window of foundation models.

\textbf{Shifted Sparse Attention Method.} \textbit{LongLoRA}~\cite{Chen2023LongLoRA} tackles the challenge by integrating position interpolation~\cite{chen2023extending} with LoRA, enabling efficient fine-tuning of LLMs for longer contexts.
Unlike standard LoRA applications, LongLoRA extends trainable low-rank adaptations to embedding and normalization layers in addition to the attention layers. A key innovation is the Shifted Sparse Attention (S$^2$-Attn) mechanism, which approximates full attention during training by partitioning the input sequence into groups and applying attention within each group. To enhance information flow between groups, half of the attention heads are shifted by half the group size. This approach facilitates efficient training on extended sequences while preserving the original model architecture during inference.

\textbf{Sink Fixed Attention Method}. Building upon LongLoRA, \textbit{SinkLoRA}~\cite{Zhang2024SinkLoRA} introduces the Sink Fixed Attention (SF-Attn) mechanism to address specific limitations. SF-Attn combines a segmentation and reassembly algorithm with global attention focused on a limited number of "sink attention tokens." This method effectively redistributes attention scores, mitigating the overemphasis on initial tokens often observed in autoregressive models. 

Another advancement, \textbit{LongQLoRA}~\cite{yang2023longqlora}, combines QLoRA~\cite{dettmers2024qlora} with position interpolation~\cite{chen2023extending} and Shifted Short Attention. By quantizing the base model to 4-bit precision during fine-tuning, LongQLoRA enables context length extension with reduced computational resources compared to LongLoRA.

These LoRA-based techniques for long sequence modeling demonstrate significant potential in extending the context window of foundation models without incurring extensive computational overhead or necessitating full model fine-tuning.

\subsection{LoRA Serving Systems}
\label{subsec:lora_serving_systems}
Efficient serving of multiple LoRA models is also essential. Recent advancements include improved GPU memory management~\cite{sheng2023s}, efficient batching techniques~\cite{chen2024punica}, CPU-assisted strategies to mitigate cold-start latency~\cite{li2024caraserve}, and adaptation methods for resource-constrained personal devices~\cite{zhaolora}.

\textbit{S-LoRA}~\cite{sheng2023s} introduced a unified paging mechanism to manage both KV cache and LoRA weights in GPU memory, enabling \textit{concurrent serving of thousands of LoRA adapters}. \textbit{Punica}~\cite{chen2024punica} developed a custom CUDA kernel, Segmented Gather Matrix-Vector Multiplication (SGMV), facilitating efficient \textit{batching of requests across different LoRA models on a single GPU}. 
\textbit{CARASERVE}~\cite{li2024caraserve} adopted a CPU-assisted strategy, initiating prefill computations for newly requested adapters during GPU loading to \textit{mitigate cold-start latency}. Furthermore, CARASERVE introduced a rank-aware scheduling algorithm for optimized request routing in multi-GPU clusters. 
\textbit{CA-LoRA}~\cite{zhaolora} incorporated LoRA knowledge inheritance and model knowledge recovery mechanisms to \textit{maintain performance on personal devices}.
These innovations enhance the scalability and efficiency of serving multiple LoRA adapters across diverse computing environments.

\section{Applications}
\label{sec:05.applications}

The effectiveness and efficiency of LoRA in fine-tuning foundation models have led to its widespread adoption across domains, including language processing, computer vision, speech recognition, multimodal learning, code engineering, scientific discovery, recommender systems, graph learning, and spatiotemporal forecasting. 

\subsection{LoRA in Language Tasks}
\label{subsec:lora_language_tasks}
Language foundation models such as LLaMA~\cite{touvron2023llama}, RoBERTa~\cite{liu2019roberta}, and DeBERTa~\cite{he2021deberta} serve as essential bases in LoRA research and have been extensively studied across tasks including natural language understanding~\cite{kopiczko2023vera,zhang2023lora, koohpayegani2023nola}, question answering~\cite{zhang2023adalora,qiang2023bilora}, machine translation~\cite{renduchintala2023tied,zhang2023lora}, reasoning~\cite{li2023loftq,renduchintala2023tied,huang2023lorahub, liu2023moelora}, and natural language generation~\cite{dettmers2024qlora,valipour2022dylora,lialin2023relora}. This section highlights LoRA's applications in specialized NLP domains.

\textbf{Multilingual Language and Dialect Processing}. 
LoRA enables efficient multilingual adaptation while preserving base models' capabilities through minimal parameter updates.
FedLFC~\cite{li2025efficient} introduces a LoRA-based fine-tuning framework that uses pretrained language-specific LoRA experts, combining them into a mixture of language experts (MoLE) and enhancing performance through knowledge distillation for efficient speech recognition.
\textbit{LAMPAT}~\cite{le2024lampat} leverages LoRA for unsupervised multilingual paraphrasing by applying adversarial training to generate diverse outputs while preserving semantic meaning across languages. 
\textbit{HyperLoRA}~\cite{xiao2023task} generates dialect-specific LoRA adapters using linguistic feature vectors, enabling zero-shot adaptation to unseen English dialects without requiring dialectal training data.

\textbf{Medical and Clinical Text Processing}. 
Medical and clinical text processing faces unique challenges due to limited data availability and the sensitive nature of clinical datasets. LoRA has emerged as a promising solution to address these limitations. Ji et al.~\cite{ji2024assertion} improved assertion detection with minimal data in clinical settings, while Le et al.~\cite{le2024impact} demonstrated LoRA's adaptability for clinical NLP tasks in data-limited environments. Liu et al.~\cite{liu2024moe} combined LoRA with MoE in a multi-task framework to tackle data imbalance. Shi et al.~\cite{shi2024medadapter} used LoRA in the MedAdapter framework for test-time adaptation, minimizing computational costs and data sharing. Christophe et al.~\cite{christophe2024med42} proposed the Med42 model, showing that LoRA fine-tuning outperforms traditional methods on key medical benchmarks such as the USMLE.

In addition, LoRA has been applied to other language tasks, including emotion understanding in conversation~\cite{zhang2023dialoguellm}, multimodal relationship extraction~\cite{li2024instruction}, and personalized text processing~\cite{zhang2024personalized}.

\subsection{LoRA in Computer Vision}
\label{subsec:lora_computer_vision}
LoRA has been effectively applied to vision foundation models such as ViTs~\cite{alexey2020image}, DINOv2~\cite{oquab2023dinov2}, MAE~\cite{he2022masked}, SAM~\cite{kirillov2023segment}, and Florence~\cite{yuan2021florence}, enhancing adaptability across visual understanding and visual generation tasks.

\subsubsection{Visual Understanding}
Visual understanding contains a broad spectrum of tasks, including domain adaptation, semantic segmentation, and content authenticity checking.

\textbf{Domain Adaptation and Transfer Learning.}
Adapting foundation models trained on extensive natural image datasets to specialized domains such as medical imaging or satellite data often presents challenges due to limited domain-specific data and computational constraints. 
To address these challenges, several studies have explored the application of LoRA for efficient domain adaptation and transfer learning in various visual tasks. 

\textbit{ExPLoRA}~\cite{khanna2024explora} integrates LoRA modules into ViTs' self-attention mechanisms, capturing domain-specific style variations, such as in satellite imagery. Similarly, \textbit{MeLo}~\cite{zhu2024melo} applies LoRA for fine-tuning in medical imaging, such as thoracic disease classification. Additionally, Kong et al.~\cite{kong2023enhancing} use LoRA to improve the generalization of ViTs for face forgery detection across various manipulation techniques and datasets.

A notable advancement in this direction is \textbit{ConvLoRA}~\cite{aleem2024convlora}, which extends the LoRA paradigm to convolutional neural networks for unsupervised domain adaptation in medical images. 
The architecture combines trainable low-rank decomposition matrices with adaptive batch normalization, establishing a more robust framework for domain transfer.

\textbf{Semantic Segmentation}. The adaptation of visual foundation models, particularly the Segment Anything Model (SAM)~\cite{kirillov2023segment}, has witnessed significant progress in semantic segmentation through LoRA integration. 
\textbit{ConvLoRA}~\cite{aleem2024convlora} enhances SAM~\cite{kirillov2023segment} for semantic segmentation in remote sensing, medical, and agricultural images.
Building upon this foundation, \textbit{SAMed}~\cite{zhang2023customized} demonstrates the effectiveness of LoRA-based fine-tuning for multi-organ segmentation tasks. 
\textbit{SurgicalSAM}~\cite{yue2024surgicalsam} applies similar techniques to the domain of robotic surgical instrument segmentation.

\textbf{Content Authenticity Checking.} Detecting synthesized content has become increasingly important with the advancement of generative models.
\textbit{CLIPMoLE}~\cite{liu2024mixture} combines shared and separate LoRA modules within an MoE framework to efficiently fine-tune CLIP ViT models for transferable image detection across diverse generation techniques. Similarly, \textbit{MoE-FFD}~\cite{kong2024moe} integrates LoRA with convolutional adapter modules, targeting face forgery detection while keeping the ViT backbone frozen.

Beyond the aforementioned applications, LoRA has also shown potential in enhancing model robustness~\cite{yuan2024fulllora}, visual tracking tasks~\cite{lin2024tracking}.

\subsubsection{Visual Generation}
By training extra small networks, LoRA enables original pretrained models, such as diffusion models~\cite{ho2020denoising,rombach2022high}, to be adapted to personalized styles and tasks or without retraining the entire foundation model. 

\textbf{Image Stylization.}
LoRA has become a key technique for image stylization, enabling rapid adaptation of diffusion models to specific artistic styles while preserving the base model's generative diversity.
\cite{shrestha2023style} employ LoRA to efficiently fine-tune Stable Diffusion~\cite{rombach2022high} for comic-style transfer, specifically targeting the style of Calvin and Hobbes comics.
Frenkel et al.~\cite{frenkel2024implicit} propose \textit{B-LoRA}, which leverages Stable Diffusion XL~\cite{podell2023sdxl} to implicitly disentangle style and content from a single image, enabling tasks such as style transfer and text-guided stylization.
Borse et al.~\cite{borse2024foura} introduce \textit{FouRA}, a frequency-domain low-rank adaptation method that mitigates distribution collapse and data copying issues common in standard LoRA fine-tuning. Recently, \textit{DragLoRA}~\cite{xia2025draglora} integrates LoRA adapters into drag-based editing to enhance precision and efficiency in foreground object manipulation.

\textbf{Multi-Concept Customization.}
LoRA is widely adopted for multi-concept customization in visual generation, where models must synthesize images combining multiple subjects or styles.
Shah et al.~\cite{Shah2023ZipLoRAAS} propose \textit{ZipLoRA}, which merges independently trained LoRAs via embedding-decomposed adaptation and gradient fusion while preserving concept identities.
Gu et al.~\cite{gu2024mix} and Zhong et al~\cite{zhong2024multi} explore LoRA switching and composition to enhance diffusion control and multi-element generation, respectively.
Yang et al.~\cite{yang2024lora} introduce \textit{LoRA-Composer}, a training-free framework that integrates multiple LoRAs through concept injection and isolation to mitigate concept vanishing and confusion.
Po et al.~\cite{po2024orthogonal} enforce orthogonal parameter updates to enable efficient merging of concept-specific LoRAs.
\textit{MoLE}~\cite{zhu2024mole} employs low-rank experts with customized datasets for human-centric generation, while Zhuang et al.~\cite{zhuang2025timestep} propose the TimeStep Master framework, assigning distinct LoRA experts to different diffusion timesteps.

LoRA has also been applied to \textbf{resolution-free generation} tasks, 
\textbit{ResAdapter} by Cheng et al.\cite{cheng2024resadapter} leverages LoRA to address the challenge of generating images at arbitrary resolutions while preserving their original style domain. Wang et al.\cite{wang2024fit} further advance this field with \textbit{FiTv2}, introducing enhanced model architecture and training strategies for resolution-free image generation.

Whether addressing multi-concept synthesis, high-resolution tasks, style transfer, or continual learning, LoRA provides a lightweight solution enabling models like Stable Diffusion and CLIP to maintain high performance while being adapted for new and complex tasks. 

\subsection{LoRA in Speech Recognition}
\label{subsec:lora_speech_recognition}
LoRA has been widely used in speech recognition tasks, particularly in efficiently fine-tuning foundation models like Wav2vec2~\cite{baevski2020wav2vec} and Whisper~\cite{radford2023whisper}.

In \textbf{fake audio detection}, Wang et al.~\cite{wang2023low} applied LoRA to fine-tune Wav2vec2~\cite{baevski2020wav2vec}, achieving performance comparable to full fine-tuning while reducing trainable parameters by a factor of 198. For \textbf{multilingual automatic speech recognition (ASR)}, Xu et al.~\cite{xu2024towards} proposed O-LoRA and O-AdaLoRA to adapt Whisper~\cite{radford2023whisper} for low-resource languages such as Uyghur and Tibetan. Similarly, Song et al.~\cite{song2024lora} introduced LoRA-Whisper, which integrates LoRA into Whisper to mitigate language interference and add new languages without degrading existing performance.
Liu et al.~\cite{yu2023low} further extended LoRA-Whisper to \textbf{low-resource ASR}, demonstrating advantages over bottleneck adapters when fine-tuning Whisper across seven low-resource languages.

\subsection{LoRA in Code Engineering}
\label{subsec:lora_code_engineering}
In the field of code engineering, LoRA has emerged as a transformative approach for enhancing processes such as code review, repair, and generation tasks. 

For \textbf{code review and analysis}, Lu et al.\cite{lu2023llama} introduced \textbit{LLaMA-Reviewer}, which fine-tunes LLaMA for review prediction, comment generation, and code refinement tasks with less than 1\% trainable parameters. Silva et al.\cite{silva2023repairllama} developed \textbit{RepairLLaMA}, which employs a lightweight repair adapter for automated program repair, enabling effective deployment in resource-constrained environments.

For \textbf{code generation and summarization}, Kumar et al.~\cite{kumar2024code} developed LoRA-based federated learning methods for code summarization, preserving data privacy without direct access to source code. Cui et al.~\cite{Cui2024OriGen} introduced \textbit{OriGen}, which leverages code-to-code augmentation and self-reflection techniques for generating high-quality register-transfer-level (RTL)\footnote{RTL is a design abstraction that describes digital circuits in terms of data flow between registers and their logical operations.} code.

\subsection{LoRA in Scientific Discovery}
\label{subsec:lora_scientific_discovery}
LoRA has been utilized across a wide range of scientific fields, including molecular tasks and materials science.

In \textbf{protein analysis}, Zeng et al.~\cite{zeng2024parameter} developed \textbit{PEFT-SP}, a framework that leverages LoRA to fine-tune large protein language models (PLMs) such as ESM-2~\cite{lin2022language} for signal peptide prediction. 
This approach significantly improved performance, especially for rare peptide types, while mitigating overfitting and maintaining low computational costs. 
Similarly, Schmirler et al.~\cite{schmirler2024fine} applied LoRA to PLMs such as ProtT5~\cite{elnaggar2021prottranscrackinglanguagelifes} and ESM-2 across diverse protein-related tasks, demonstrating accelerated training and improved downstream predictions while preventing catastrophic forgetting.
In another molecular application, Schreiber~\cite{schreiber2023esmbind} employed LoRA and its quantized variant, QLoRA~\cite{dettmers2024qlora}, in the ESMBind and QBind models for protein binding-site and post-translational modification prediction. 
These models achieved enhanced generalization on unseen protein sequences without relying on structural data or multiple sequence alignments.
Lv et al.~\cite{lv2024prollama} introduced \textbit{ProLLaMA}, demonstrating that a two-stage LoRA approach with varying ranks enables effective protein language learning while maintaining natural language capabilities.

Beyond molecular science, Buehler et al.~\cite{buehler2024x} developed \textbit{X-LoRA}, a framework that dynamically combines specialized LoRA adapters to solve inverse and forward problems in protein mechanics and material design. Dagdelen et al.~\cite{dagdelen2024structured} utilized LoRA to enhance scientific information extraction, enabling more effective data processing and knowledge discovery. These methods highlight LoRA's potential for complex multidisciplinary applications.

\subsection{LoRA in Recommender Systems}
\label{subsec:lora_recommender_systems}
In recommender systems, LoRA efficiently fine-tunes LLMs for CTR prediction and sequential recommendation tasks. 

For \textbf{CTR prediction}, Yang et al.\cite{yang2024mlora} proposed \textbit{MLoRA}, implementing domain-specific low-rank matrices to capture inter-domain variations and enhance personalization, successfully deployed at Alibaba. Zhu et al.\cite{zhu2024lifelong} introduced \textbit{RecLoRA}, which replaces single LoRA matrices with a meta-LoRA structure, using soft routing to select personalized matrix combinations for each user's evolving interests.

For \textbf{sequential recommendation}, Qin et al.~\cite{qin2024atflrec} introduced \textbit{ATFLRec}, which integrates audio and text data by independently optimizing LoRA modules for each modality. Zheng et al.~\cite{zheng2024harnessing} developed \textbit{LLM-TRSR} for text-rich recommendation scenarios, leveraging LoRA's efficient fine-tuning to handle large-scale text data while maintaining real-time responsiveness. Kong et al.~\cite{kong2024customizing} proposed \textbit{iLoRA}, which employs an MoE framework with specialized LoRA modules for distinct user behavior patterns, dynamically engaging relevant experts based on individual interaction sequences. 
Recently, Ji et al.~\cite{ji2024genrec} introduced \textbit{GenRec}, a text-centric LLM that treats item names as identifiers and fine-tunes LLaMA with LoRA, achieving superior performance on large-scale datasets by directly learning collaborative information from textual representations.

\subsection{LoRA in Graph Learning}
\label{subsec:lora_graph_learning}
Recent work has also explored applying LoRA to non-Euclidean data (i.e., graphs) by fine-tuning graph neural networks (GNNs)~\cite{sun2022gppt} to adapt to new graphs or structural updates of existing ones.

For \textbf{cross-domain graph neural network adaptation}, Yang et al.~\cite{yang2024graphlora} introduced \textbit{GraphLoRA}, which constructs a small trainable GNN alongside the pretrained model to bridge structural and feature distribution gaps between graphs. By applying low-rank decomposition to the tunable GNN parameters and incorporating graph-structure-based regularization, GraphLoRA adapts pretrained GNNs to diverse domains while fine-tuning only 20\% of the parameters.

For \textbf{dynamic knowledge graph learning}, Liu et al.\cite{liu2024fast} developed \textbit{IncLoRA}, which adapts entity and relation embeddings to continuous graph updates. IncLoRA groups new knowledge embeddings into explicit LoRA layers based on their distance to preserved graphs and allocates rank scales adaptively using graph structural properties.

For \textbf{equivariant GNNs}, ELoRA~\cite{wang2025elora} is tailored to SO(3)-equivariant architectures, maintaining physical symmetry during fine-tuning and improving the accuracy of molecular energy and force predictions.

\subsection{LoRA in Spatial-Temporal Forecasting}
\label{subsec:lora_spatial-temporal_forecasting}
Multivariate time series data is prevalent in real-world scenarios such as transportation, weather forecasting, and economics~\cite{lim2021time, zhang2025adapting, Pan2025MixtureOL}. Recent studies have explored using LoRA to address challenges in this field, including node-specific adaptation, multi-channel modeling, and out-of-domain prediction.

For \textbf{node-specific adaptation}, Ruan et al.\cite{ruan2024low} proposed \textbit{ST-LoRA}, which implements node-adaptive LoRA layers to add extra learnable parameters for each node. By incorporating residual structure among LoRA layers to avoid overparameterization, ST-LoRA effectively captures distinct patterns and dynamics of different nodes in transportation datasets.

For \textbf{multi-channel modeling}, Nie et al.\cite{nie2024channel} introduced \textbit{C-LoRA}, which balances between channel-dependent and channel-independent strategies. C-LoRA parameterizes each channel with a low-rank factorized adapter to form identity-aware embeddings, then inputs these to a globally shared predictor for modeling cross-channel dependencies.

For \textbf{out-of-domain prediction}, Gupta et al.~\cite{gupta2024low} analyzed the effectiveness of LoRA-based fine-tuning across leading time-series foundation models such as Lag-Llama~\cite{rasul2023lag}, MOIRAI~\cite{woo2024unified}, and Chronos~\cite{ansari2024chronos}, demonstrating improved forecasting of sepsis patients' vital signs in intensive care units while reducing computational costs. 
Ren et al.\cite{ren2024tpllm} introduced \textbit{TPLLM}, which injects trainable rank-decomposition matrices into GPT-2 Transformer blocks for traffic forecasting, effectively adapting the model to process spatial-temporal representations with limited historical traffic data.

\subsection{LoRA in Multi-Modal}
\label{subsec:multi_modal}
Multimodal foundation models (MFMs) combine modalities such as text, audio, images, and video within a shared representational space, enabling cross-modal reasoning and understanding. LoRA enhances these models by improving training efficiency while strengthening inter-modal alignment. Language-vision and language-audio learning represent two primary domains where LoRA has been extensively applied to adapt MFMs.

\subsubsection{Language-vision Learning}
In \textbf{Language-vision learning tasks}, LoRA has been applied particularly in enhancing vision-language capabilities and customizing diffusion models.

\textit{Language-Vision Model-Based Adaptation.} 
Sung et al.~\cite{sung2022vl} adapted vision-language models (VLMs) by fine-tuning CLIP-BART for visual question answering and image captioning.
Ji et al.~\cite{ji2024advlora} enhanced the adversarial robustness of VLMs through clustering-based LoRA for cross-modal retrieval. Zong et al.~\cite{zong2024mova} proposed \textbit{MoVA}, which efficiently routes and fuses multiple vision foundation models such as CLIP~\cite{radford2021learning}, DINOv2~\cite{oquab2023dinov2}, and SAM~\cite{kirillov2023segment} through a coarse-to-fine adapter mechanism. In addition, \textbit{D-MoLE}~\cite{ge2025dynamic} employs a layer-wise expert allocator to assign LoRA experts automatically across layers, resolving architectural conflicts and routing instructions to facilitate knowledge sharing among experts, thereby improving language-vision performance.

\textit{Diffusion Model Based Customization.} 
Various approaches have leveraged LoRA to efficiently adapt Stable Diffusion (SD) models. 
For concept customization, Kumari et al.\cite{kumari2023multi} introduced \textit{Custom Diffusion} with LoRA in domain adapter layers for few-shot concept learning, while Li et al.\cite{li2024selma} developed \textbit{SELMA} to train and merge multiple skill-specific LoRA experts without interference. 
Lu et al.\cite{lu2024mace} advanced concept manipulation through LoRA modules with cross-attention refinement for selective concept erasure. 
Additional applications include \textbit{StitchDiffusion}~\cite{wang2024customizing} for 360-degree panorama generation, Chinese garden image synthesis~\cite{shi2023space}, \textbit{DreamSync}~\cite{sun2023dreamsync} for improved generation faithfulness, \textbit{Block-wise LoRA}~\cite{li2024blockwise} for fine-grained control, and \textbit{AnimateDiff}~\cite{guo2023animatediff} for motion adaptation.

In vision-language model adaptation, LoRA has enhanced cross-modal understanding through techniques like CLIP-BART fine-tuning~\cite{sung2022vl}, clustering-based approaches~\cite{ji2024advlora}, and multi-model fusion~\cite{zong2024mova}. For diffusion model customization, LoRA has enabled efficient concept learning~\cite{kumari2023multi}, selective manipulation~\cite{lu2024mace}, and specialized generation tasks ranging from panorama creation~\cite{wang2024customizing} to motion adaptation~\cite{guo2023animatediff}.

\subsubsection{Language-Audio Learning}
In \textbf{language-audio learning tasks}, LoRA has emerged as a pivotal technique for addressing the modality gap between audio and text representations in foundation models. Applications primarily fall into two categories: speech recognition and audio content understanding and generation.

\textit{Speech Recognition.}
Fathullah et al.~\cite{fathullah2024prompting} introduced a direct audio-embedding method in which LoRA adapts self-attention layers for audio-text alignment, enabling large language models to process audio inputs effectively. 
Yusuf et al.~\cite{yusuf2024speculative} proposed speculative speech recognition by combining an RNN-Transducer-based ASR system~\cite{graves2012sequence} with an audio-prefixed language model adapted via LoRA, allowing the model to generate speculative predictions.
Palaskar et al.~\cite{palaskar2024multimodal} further introduced FLoRA with modality-specific adapters that can be selectively engaged, providing a robust framework for multimodal integration. 

\textit{Speech Content Understanding and Generation.}
\textbit{LOAE}~\cite{liu2024enhancing} combines CED-based audio encoding~\cite{dinkel2024ced} with LLaMA-based text decoding for automated audio captioning, using a Q-Former~\cite{li2023blip} bridge and LoRA for efficient optimization. 
Qin et al.~\cite{qin2024atflrec} introduced ATFLRec, a multimodal recommender system that leverages separate LoRA modules to fine-tune audio and text modalities in LLMs, demonstrating superior performance through a fusion-based approach.

These developments in language-audio learning have established several key principles for applying LoRA in multi-modal contexts: (1) the importance of modality-specific adaptation paths~\cite{fathullah2024prompting,palaskar2024multimodal}, (2) the effectiveness of selective parameter updating for cross-modal alignment~\cite{yusuf2024speculative,liu2024enhancing}, and (3) the value of maintaining distinct representational spaces through specialized LoRA modules before fusion~\cite{palaskar2024multimodal,qin2024atflrec}.

\section{Challenges and Discussion}
\label{sec:06.discussion}
While LoRA has excelled in efficiently adapting foundation models across diverse domains, several critical challenges and opportunities require further investigation.

\textbf{Theoretical Understanding.} Existing theoretical frameworks primarily focus on simplified settings or specific architectures~\cite{zeng2024the}, leaving gaps in our understanding of LoRA's behavior in more complex scenarios. For instance, the interaction between LoRA adaptations and the pretrained model's knowledge is not fully understood, particularly in how LoRA preserves useful features while modifying task-specific ones. Developing more comprehensive theoretical frameworks that can explain LoRA's effectiveness across different architectures and tasks would be valuable for guiding future improvements.

\textbf{Architectural Design Principles.} Current LoRA implementations often rely on empirical observations rather than systematic design methodologies~\cite{hu2023llm,hu2021lora}. Critical questions persist regarding optimal adapter placement strategies, rank determination across network depths, and the geometric properties of adaptation spaces. Recent explorations into non-Euclidean geometries, such as hyperbolic spaces~\cite{yang2024hyperbolic}, suggest potential advantages for capturing hierarchical relationships in model adaptations. A unified framework for analyzing these design choices could significantly advance our understanding of the parameter efficiency versus model capacity trade-off.

\textbf{Computational Efficiency.} The scalability of LoRA becomes increasingly crucial as LLMs continue to expand. Managing concurrent adaptations while handling variable-length sequences presents significant challenges in memory usage and computational overhead. The dynamic management of adapter modules and KV cache tensors can lead to memory fragmentation and increased I/O costs~\cite{sheng2023s}. Advanced serving architectures and optimization techniques are essential for maintaining low latency in production environments, particularly when dealing with multiple concurrent adaptation requests.

\textbf{Robustness and Verification.} The deployment of LoRA-adapted models in critical applications necessitates robust verification methods. Current research inadequately addresses model behavior under distribution shifts and adversarial conditions~\cite{wang2023dala}. Developing rigorous uncertainty quantification methods and formal verification techniques becomes paramount, especially for high-stakes applications in healthcare and autonomous systems, where model reliability directly impacts human safety.

\textbf{Privacy and Security.} As LoRA becomes more widely adopted, particularly in federated learning settings, privacy and security considerations become increasingly important. This includes protecting sensitive information in LoRA adaptations, preventing unauthorized access or manipulation of adapted models, and ensuring that LoRA updates do not inadvertently leak private information~\cite{sun2024improving,zhu2024promoting}. Research into privacy-preserving LoRA adaptation techniques and secure methods for sharing and combining LoRA modules is needed.

Future research directions should focus on establishing comprehensive theoretical frameworks that unify various LoRA design aspects; developing automated architecture search methods for optimal adapter configurations; creating efficient serving infrastructures for large-scale deployments; implementing standardized evaluation protocols for reliability assessment; and integrating advanced privacy-preserving mechanisms into the adaptation process.

Moreover, as novel architectures emerge, such as Mamba~\cite{halloran2024mamba}, investigating LoRA's applicability to these new paradigms becomes crucial. The integration of LoRA into edge computing and real-time systems~\cite{jihong2024edge} presents additional challenges that intersect with hardware optimization and system design, necessitating interdisciplinary research efforts.

\section{Conclusion}
\label{sec:07.conclusion}
In this survey, we have presented a systematic analysis of LoRA, examining its theoretical underpinnings, technical advances, and diverse applications in adapting foundation models. The extensive adoption of LoRA across diverse domains—from natural language processing and computer vision to speech recognition and scientific computing—highlights its versatility and effectiveness. Its success in maintaining model performance while significantly reducing computational and storage requirements has made it particularly valuable for resource-constrained environments and specialized domain adaptations.
Despite these achievements, several critical challenges persist. The theoretical framework underlying LoRA’s effectiveness requires further development, particularly in understanding the interaction between low-rank adaptations and model capabilities. Additionally, questions regarding scalability, robustness, and secure deployment in production environments present ongoing research opportunities.

\ifCLASSOPTIONcaptionsoff
  \newpage
\fi

\bibliographystyle{IEEEtran}
\bibliography{references}

\end{document}